%% file: example_paper.tex
\documentclass{article}

\input{math_commands.tex}

\usepackage{microtype}
\usepackage{graphicx}
\usepackage{subcaption}
\usepackage{booktabs} % for professional tables

\usepackage{threeparttable}
\usepackage{url}
\usepackage{siunitx}
\usepackage{multirow}
\usepackage{xcolor,colortbl}
\usepackage[most]{tcolorbox}
\usepackage{tabularx}
\usepackage{xcolor}
\usepackage{verbatim}
\usepackage{pifont}
\usepackage{algorithm}
\usepackage{algpseudocode}
\usepackage[normalem]{ulem}

\newcommand{\xmark}{\ding{55}}
\usepackage{capt-of}
\usepackage{wrapfig}
\usepackage{placeins}

\usepackage{hyperref}

\usepackage[accepted]{icml2026}

\usepackage{amsmath}
\usepackage{amssymb}
\usepackage{mathtools}
\usepackage{amsthm}

\usepackage[capitalize,noabbrev]{cleveref}

\theoremstyle{plain}

\theoremstyle{definition}

\theoremstyle{remark}

\usepackage[textsize=tiny]{todonotes}

\icmltitlerunning{Evolution Strategies at Scale: LLM Fine-Tuning Beyond Reinforcement Learning}

\begin{document}

\twocolumn[
  \icmltitle{Evolution Strategies at Scale: \\LLM Fine-Tuning Beyond Reinforcement Learning}

  % It is OKAY to include author information, even for blind submissions: the
  % style file will automatically remove it for you unless you've provided
  % the [accepted] option to the icml2026 package.

  % List of affiliations: The first argument should be a (short) identifier you
  % will use later to specify author affiliations Academic affiliations
  % should list Department, University, City, Region, Country Industry
  % affiliations should list Company, City, Region, Country

  % You can specify symbols, otherwise they are numbered in order. Ideally, you
  % should not use this facility. Affiliations will be numbered in order of
  % appearance and this is the preferred way.
  \icmlsetsymbol{equal}{*}
  \icmlsetsymbol{lead}{\textdagger}
  \icmlsetsymbol{intern}{\textdaggerdbl}

  \begin{icmlauthorlist}
    \icmlauthor{Xin Qiu}{equal,lead,cognizant}
    \icmlauthor{Yulu Gan}{equal,intern,mit}
    \icmlauthor{Conor F. Hayes}{equal,cognizant}
    \icmlauthor{Qiyao Liang}{intern,mit}
    \icmlauthor{Yinggan Xu}{intern,ucla}
    \icmlauthor{Roberto Dailey}{cognizant}\\
    \icmlauthor{Elliot Meyerson}{cognizant}
    %\icmlauthor{}{sch}
    \icmlauthor{Babak Hodjat}{cognizant}
    \icmlauthor{Risto Miikkulainen}{cognizant,ut}
    %\icmlauthor{}{sch}
    %\icmlauthor{}{sch}
  \end{icmlauthorlist}

  \icmlaffiliation{ucla}{University of California, Los Angeles, Los Angeles, CA, USA}
  \icmlaffiliation{cognizant}{Cognizant AI Lab, San Francisco, CA, USA}
  \icmlaffiliation{ut}{The University of Texas at Austin, Austin, TX, USA}
  \icmlaffiliation{mit}{Massachusetts Institute of Technology, Cambridge, MA, USA}

  \icmlcorrespondingauthor{Xin Qiu}{qiuxin.nju@gmail.com}

  % You may provide any keywords that you find helpful for describing your
  % paper; these are used to populate the "keywords" metadata in the PDF but
  % will not be shown in the document
  \icmlkeywords{Machine Learning, ICML}

  \vskip 0.3in
]

% this must go after the closing bracket ] following \twocolumn[ ...

% This command actually creates the footnote in the first column listing the
% affiliations and the copyright notice. The command takes one argument, which
% is text to display at the start of the footnote. The \icmlEqualContribution
% command is standard text for equal contribution. Remove it (just {}) if you
% do not need this facility.

% Use ONE of the following lines. DO NOT remove the command.
% If you have no special notice, KEEP empty braces:
\printAffiliationsAndNotice{\icmlEqualContribution. \textsuperscript{\textdagger}Project Lead. \textsuperscript{\textdaggerdbl}Work done during an internship at Cognizant AI Lab.}  % no special notice (required even if empty)
% Or, if applicable, use the standard equal contribution text:
% \printAffiliationsAndNotice{\icmlEqualContribution}

\begin{abstract}

Fine-tuning large language models (LLMs) for downstream tasks is an essential stage of modern AI deployment. Reinforcement learning (RL) has emerged as the dominant fine-tuning paradigm, underpinning many state-of-the-art LLMs. In contrast, evolution strategies (ES) has largely been overlooked due to the widespread belief that it does not scale to modern model sizes. This paper overturns this assumption by demonstrating the first successful application of ES to full-parameter fine-tuning of LLMs at the billion-parameter scale, without dimensionality reduction. ES can indeed search over extremely high-dimensional parameter spaces and outperform established RL implementations across multiple axes, including
% the elimination of backpropagation,
improved tolerance to long-horizon and delayed rewards, robustness across diverse base LLMs, reduced susceptibility to reward hacking, and improved training stability. These findings suggest that ES is not merely a viable alternative to RL, but a fundamentally different and powerful backpropagation-free post-training paradigm that opens a new direction for LLM fine-tuning beyond current RL-based approaches.

%   Fine-tuning large language models (LLMs) for down-stream tasks is a critical step in the AI deployment pipeline. Reinforcement learning (RL) is arguably the most prominent fine-tuning method, contributing to the birth of many state-of-the-art LLMs. In contrast, evolution strategies (ES), which once showed comparable performance to RL on models with a few million parameters, was neglected due to the pessimistic perception of its scalability to larger models. In this work, we report the first successful attempt to scale up ES for fine-tuning the full parameters of LLMs, without dimension reduction, showing the surprising fact that ES can search efficiently over billions of parameters and existing RL baselines in multiple respects, including backpropagation-free, tolerance to long-horizon rewards, robustness to different base LLMs, less tendency to reward hacking, and more stable performance across runs. It therefore serves as a basis to unlock a new direction in LLM fine-tuning beyond what current RL techniques provide. %\conor{I do not want to touch the abstract directly, so please see my comment in the overleaf comments.}
\end{abstract}

\section{Introduction} \label{sec:intro}
As the capabilities of large language models (LLMs) have rapidly improved, these systems have been increasingly deployed across scientific and engineering workflows \citep{touvron2023llama2,openai2024gpt4,llama3modelcard,jiang2024mixtral,liu2024deepseek,claude4,gemini25,singhal2023expertlevel,wu2023bloomberggpt,roziere2024code,Romera-Paredes2024math}. This widespread deployment has made fine-tuning a standard step for adapting pre-trained models to downstream tasks and aligning behavior with user preferences \citep{ouyang2022rlhf,rafailov2023dpo,latif2024score, deepseekai2025deepseekr1}. In practice, reinforcement learning (RL) has become the predominant choice for this fine-tuning stage \citep{ouyang2022rlhf, bai2022assistant, shao2024grpo, deepseekai2025deepseekr1,liang2025deepseeknature,srivastava2025rlsurvey}. However, several challenges have emerged: First, RL methods incur low sample efficiency and high variance of the gradient estimator when handling long-horizon rewards, which is a common case for LLM fine-tuning with outcome-only rewards \citep{salimans2017es, sutton2018reinforcement,vemula2019contrast}. Proper credit assignment at token level for RL fine-tuning methods is difficult and possibly unhelpful \citep{zhang2025lessons,song2025prmbench,liang2025deepseeknature,uesato2022math,jia2025need,liang2025deepseeknature}. Second, RL techniques are sensitive to the choice of base LLMs, resulting in inconsistent fine-tuning performance across different models \citep{gandhi2025cognitive}. Third, RL techniques tend to incentivize hacking the reward function, leading to undesirable behaviors \citep{gao2023scaling,denison2024tampering,fu2025shaping}. Fourth, RL is sensitive to hyperparameters and often unstable across multiple runs even with the same hyperparameter setup, significantly increasing fine-tuning cost \citep{choshen2020instablerl,zhong2025dpo}. \renewcommand{\thefootnote}{}
\footnotetext{\textbf{Source code}: \href{https://github.com/VsonicV/es-at-scale}{https://github.com/VsonicV/es-at-scale}\vspace{-5pt}}
\renewcommand{\thefootnote}{\arabic{footnote}} 

Evolution Strategies (ES), a class of population-based zeroth-order optimization algorithms, is a possible alternative. ES has several advantages over RL in traditional control and gaming problems: it parallelizes naturally, tolerates long-horizon rewards, promotes broad exploration, avoids expensive backpropagation, and remains robust across hyperparameter settings \citep{salimans2017es,chrabaszcz2018,conti2018}. However, ES remains relatively underexplored in LLM fine-tuning settings. Standard ES directly optimizes in the full parameter space, which in prior applications typically contained no more than a few million parameters \citep{salimans2017es,zhang2017essgd,lehman2018esmore,lorenc2025estransformer}. It was assumed that for very large models, exploration in parameter space is significantly more difficult and sample-inefficient than exploration in action space \citep{vemula2019contrast}. Modern LLMs typically contain billions of parameters, which makes direct ES optimization appear infeasible. Existing workarounds include restricting ES to the final layer of the base model \citep{toledano2022mt5}, applying ES to low-dimensional adapters \citep{jin2024lowrank}, and performing evolutionary search in action space, analogous to standard RL \citep{huang2025evolution}. Directly searching in the full parameter space of LLMs (without dimensionality reduction) has remained a challenge.

This paper is aimed at meeting this challenge. For the first time, ES is scaled to multi-billion-parameter search spaces through direct optimization of the full parameter space of LLMs during fine-tuning. The approach is based on a memory-efficient implementation of an algorithmically simplified ES variant, with support for parallelization across GPUs. Performance is compared with state-of-the-art (SOTA) RL methods in fine-tuning various LLMs in several reasoning benchmark tasks, and behavioral differences from RL are analyzed in terms of fine-tuning for conciseness. Furthermore, ES fine-tuning is successfully applied to solve two puzzle problems that are challenging for base LLMs. 

ES was found able to search directly over billions of parameters without dimensionality reduction while achieving strong fine-tuning performance relative to RL in multiple aspects: (1) \emph{ES only needs response-level rewards}, making it a perfect fit for fine-tuning on reasoning tasks that have only sparse long-horizon outcome rewards. In particular, ES obtained significantly better fine-tuned models than RL in the Countdown task with such rewards. (2) \emph{ES is able to find good solutions in large space with small populations}, e.g.\ just 30 in the multi-billion-parameter space in this paper.
As a comparison, previous ES implementations \citep{salimans2017es,zhang2017essgd,lehman2018esmore,lorenc2025estransformer} utilized a population size of 10,000 or more with much smaller models (i.e.\ millions of parameters or less). The current extremely small population size thus makes the approach feasible even without extensive compute.
(3) \emph{ES is more robust} than RL across different LLMs. While RL fine-tuning failed on some LLMs, ES provided good fine-tuning for all of them. ES benefits from its exploration in parameter space, making it less sensitive to initial states of the LLMs.
(4) \emph{ES consistently maintains reasonable behaviors} during fine-tuning, in contrast to RL that tends to hack the reward function if no other penalty is added. The main reason is that ES optimizes a solution distribution \citep{lehman2018esmore}, which is more difficult to hack, while RL optimizes a single solution.
(5) \emph{ES's behavior is more consistent} than RL's across different runs. This property can significantly reduce expected cost of fine-tuning.
(6) \emph{Fine-tuning with ES only requires inference}, and therefore no gradient calculations are needed. A significant amount of GPU memory can therefore be saved. %\citep{malladi2023mezo}.

% The study reported in this paper serves as a first step in demonstrating the potential of ES for fine-tuning LLMs. The surprising and counterintuitive findings motivates further work scaling up ES to larger LLM fine-tuning tasks. Given the unique advantages over the state of the art, ES opens up new opportunities in parameter-space exploration, outcome-only fine-tuning, and large-scale distributed post-training.

Thus, this study establishes a critical first milestone in demonstrating that ES can serve as a viable and powerful post-training paradigm for LLMs. The results reveal a surprising and counterintuitive finding that ES remains effective when scaled to models with billions of parameters, directly challenging the long-held assumption that such methods are inherently unscalable. These findings not only motivate further scaling to even larger LLMs, but fundamentally expand the design space of post-training algorithms. By operating directly in parameter space without reliance on gradients or intermediate supervision, ES enables new forms of outcome-only optimization, robust exploration over high-dimensional parameter landscapes, and naturally distributed large-scale fine-tuning. Taken together, this paper positions ES as a foundational alternative to gradient-based RL and opens a new direction for scalable, stable, and general LLM post-training.

\section{Related Work}

The background on Evolution Strategies and evolutionary optimization of LLMs is first reviewed, followed by SOTA RL fine tuning and parameter-space exploration. 

\textbf{Traditional ES:} Evolution Strategies \citep[ES,][]
{rechenberg1973es,schwefel1977es} are a class of evolutionary algorithms (EAs) for solving numerical optimization problems. The main idea is to sample a population of solutions through perturbations, then recombine the perturbed solutions based on their fitness values to form the population for the next generation. This process repeats until a termination condition is triggered, e.g., the maximum number of generations is reached. Among the different variants of ES, CMA-ES \citep{hansen2001cmaes}, which utilizes a multivariate Gaussian distribution with full covariance matrix to sample the population, and natural ES \citep{wierstra2008,wierstra14a}, which uses natural gradient to guide the search, are two popular methods for traditional optimization problems. Although ES has long been used to evolve parameters of neural networks (NNs), \citep{igel:cec03}, \citet{salimans2017es} were the first to scale the approach up to deep learning networks. Comparable performance to RL methods in control and gaming environments was observed, and several unique advantages of ES highlighted. This seminal work paved the way for several follow-up studies. \citet{zhang2017essgd} used ES to optimize a convolutional NN with around three million parameters. They found that with a large enough population size, ES can approximate the performance of traditional stochastic gradient descent (SGD). \citet{lehman2018esmore} further optimized a NN comprising nearly 167,000 parameters with both ES and a finite-difference (FD) gradient estimator. Because ES optimizes the average reward for the entire population, whereas FD optimizes the reward for a single solution, it obtained models that were more robust to parameter perturbations. \citet{lorenc2025estransformer} applied ES to optimize decision transformers in RL environments, and observed promising results for model sizes up to around 2.5 million parameters. In a related study, another traditional EA, namely genetic algorithm (GA) with mutations only, was extended to a high-dimensional space \citep{such2017DeepNG}. Encouraging results were observed in different types of models with up to around four million parameters \citep{such2017DeepNG, risi2019gaworld}. However, although these studies were promising, the scale of these implementations was still significantly less than the size of current LLMs.

\textbf{Evolution+LLMs:} Synergies between Evolutionary Algorithms (EAs) and LLMs have received increasing attention in recent years \citep{wang2025eallm,xingyu2025eallm}. Popular research directions include EAs for prompt optimization \citep{sun2022bbt,sun2022bbtv2,zhao2023gaprompt,guo2024connecting}, utilizing LLMs as evolutionary operators \citep{meyerson2024lmx,lehman2024evolm,Romera-Paredes2024math,novikov2025alphaevolve}, and merging LLMs through evolution \citep{du2024fusion,akiba2025merge}. Applying EAs to optimize billions of parameters in LLMs is generally perceived to be intractable, but a few studies have been successful at a smaller scale. For example, \citet{toledano2022mt5} fine-tuned the last layer (with 325 parameters) of an mT5-based transformer via CMA-ES. \citet{jin2024lowrank} optimized the low-rank adapter parameters (with dimensionality up to 1600) using CMA-ES and the Fireworks algorithm. \citet{sanchez2024gatransformer} applied a GA to fine-tune around 9.5 million parameters of a transformer encoder, though poorer performance than the traditional Adam optimizer was observed. \citet{huang2025evolution} proposed a hybrid algorithm that performs exploration in action space instead of parameter space, and it was only used in the final epoch of supervised fine-tuning (SFT). The work in this paper significantly extends this prior research by successfully scaling ES to search in the billions of parameters of LLMs, leading to surprisingly good fine-tuning performance.

\textbf{RL for fine-tuning:} Fine-tuning using RL is a critical step during the training of many landmark LLMs \citep{ouyang2022rlhf, bai2022assistant, shao2024grpo, deepseekai2025deepseekr1,liang2025deepseeknature}. Proximal Policy Optimization \citep[PPO;][]{schulman2017ppo} and Group Relative Policy Optimization \citep[GRPO;][]{shao2024grpo} are the two predominant methods. PPO introduces a clipped surrogate objective to limit the update scale in each step with respect to the old policy, and it usually works with a value model in an actor-critic manner. GRPO simplifies the pipeline of PPO by replacing the value model with group advantage, which is calculated based on direct evaluations of multiple responses. As discussed in Section~\ref{sec:intro}, in the context of LLM fine-tuning, these methods struggle with several fundamental limitations, including the dilemma in handling long-horizon reward \citep{vemula2019contrast,salimans2017es,zhang2025lessons,song2025prmbench,uesato2022math,jia2025need,liang2025deepseeknature}, sensitivity to base LLMs \citep{gandhi2025cognitive}, tendency to hack reward \citep{gao2023scaling,denison2024tampering,fu2025shaping}, and instability across runs \citep{choshen2020instablerl,zhong2025dpo}. ES inherently avoids these limitations, leading to better fine-tuning performance.

\textbf{Parameter-space exploration:} Existing RL fine-tuning methods are overwhelmingly based on action-space exploration.
Parameter space exploration has received much less attention, though some such studies do exist \citep{thomas2008state,sehnke2010parameter,thomas2010parameter,plappert2018parameter}. Although promising performance was observed in problems with sparse rewards, the scale of the tested models was far smaller than that of LLMs. \citet{vemula2019contrast} performed a theoretical analysis of different exploration strategies, and found that the complexity of the parameter space exploration increased quadratically with the number of parameters, whereas the complexity of action space exploration depended on action dimensionality quadratically and horizon length of the reward quartically. Based on the classical SPSA method \citep{spall1992spsa}, \citet{malladi2023mezo} proposed a zeroth-order optimizer MeZO that directly worked in parameter space for fine-tuning LLMs. MeZO significantly reduced memory requirements, but its fine-tuning performance was no better than other baselines. In contrast, the ES implementation in this paper performs exploration in multi-billion-parameter search spaces, and exhibits strong performance across different benchmarks.

\section{Method}
This section introduces the basic ES algorithm and a detailed description of its implementation for LLM fine-tuning.
\subsection{Basic ES algorithm}
\begin{algorithm}[h]
	\caption{Basic ES Algorithm}
	\label{alg:es-llm_general}
	\begin{algorithmic}[1]
		\Require Pretrained LLM with initial parameters $\bm{\theta}_0$, reward function $R(\cdot)$, total iterations $T$, population size $N$, noise scale $\sigma$, learning rate $\alpha$.
		\For{$t = 1$ to $T$} \Comment{outer ES iterations}
		\For{$n = 1$ to $N$}
		\State Sample noise $\bm{\varepsilon}_n \sim \mathcal{N}(0,\bm{I})$
		\State Compute reward for perturbed parameters: \Statex\hspace{\algorithmicindent}\hspace{\algorithmicindent}$R_n\leftarrow R(\bm{\theta}_{t-1}+\sigma \cdot \bm{\varepsilon}_n)$
		\EndFor
		\State Normalize $R_n$
		\State Update model parameters as 
        \Statex\hspace{\algorithmicindent}$\bm{\theta}_{t} \leftarrow \bm{\theta}_{t-1}+ \alpha \cdot \frac{1}{N}\sum_{n=1}^N R_n \bm{\varepsilon}_n$
		\EndFor
	\end{algorithmic}
\end{algorithm}
\begin{algorithm*}[t]
	\caption{ES Implementation for LLM Fine-Tuning}
	\label{alg:es-llm_specific}
	\begin{algorithmic}[1]
		\Require Pretrained LLM with initial parameters $\bm{\theta}_0$, reward function $R(\cdot)$, total iterations $T$, population size $N$, noise scale $\sigma$, learning rate $\alpha$, number of parallel process $P$.
		\State Create $P$ processes, each instantiates a model with the same initial parameters $\bm{\theta}_0$, with one process as the main process
		\For{$t = 1$ to $T$} \Comment{ES iterations}
		\State Sample N random seeds $s_1, s_2, \ldots, s_N$
		\State Assign random seeds to $P$ processes
		\For{$n = 1$ to $N$}
		\State For the process handling $s_n$, reset its random number generator using random seed $s_n$
		\For{\textbf{each LLM layer}} \Comment{perturbation within current process}
		\State Sample noise $\bm{\varepsilon}_{n,l} \sim \mathcal{N}(0,\bm{I})$, which has the same shape as the $l$th layer's parameters
		\State Perturb the $l$th layer's parameters in-place: $\bm{\theta}_{t-1,l}\leftarrow\bm{\theta}_{t-1,l}+\sigma \cdot \bm{\varepsilon}_{n,l}$
		\EndFor
		\State Compute reward for perturbed parameters $R_{n}\leftarrow R(\bm{\theta}_{t-1})$ \Comment{within current process}
		\State For the process handling $s_n$, reset its random number generator using random seed $s_n$
		\For{\textbf{each LLM layer}} \Comment{restoration within current process}
		\State Sample noise $\bm{\varepsilon}_{n,l} \sim \mathcal{N}(0,\bm{I})$, which has the same shape as the $l$th layer's parameters
		\State Restore the $l$th layer's parameters in-place: $\bm{\theta}_{t-1,l}\leftarrow\bm{\theta}_{t-1,l}-\sigma \cdot \bm{\varepsilon}_{n,l}$
		\EndFor
		\EndFor
		\State Normalize the reward scores by calculating the $z$-score for each $R_n$: $Z_n\leftarrow\frac{R_n-R_{\mathrm{mean}}}{R_{\mathrm{std}}}$, 
		\Statex\hspace{\algorithmicindent}where $R_{\mathrm{mean}}$ and $R_{\mathrm{std}}$ are the mean and standard deviation of $R_1, R_2, \ldots, R_N$.
		\For{$n = 1$ to $N$} \Comment{in main process only}
		\State Reset current random number generator using random seed $s_n$
		\For{\textbf{each LLM layer}}
		\State Sample noise $\bm{\varepsilon}_{n,l} \sim \mathcal{N}(0,\bm{I})$, which has the same shape as the $l$th layer's parameters
		\State Update $l$th layer's parameters in-place as $\bm{\theta}_{t,l} \leftarrow \bm{\theta}_{t-1,l}+ \alpha \cdot \frac{1}{N} Z_n \bm{\varepsilon}_{n,l}$
		\EndFor
		\EndFor
		\State Update the model parameters of all processes to $\bm{\theta}_{t}$
		\EndFor
	\end{algorithmic}
\end{algorithm*}
The ES implementation used in this paper is based on a simplified variant of Natural Evolution Strategies (NES) \citep{wierstra2008,wierstra14a} and follows the design of OpenAI ES \citep{salimans2017es}, which employs fixed-covariance perturbation noise.

Given a pretrained LLM with initial parameters $\bm{\theta}_0$ and a target reward function $R(\cdot)$, the task is to fine-tune the parameters so that the reward function is optimized (Algorithm~\ref{alg:es-llm_general}). In each iteration, $N$ perturbed models are sampled by adding random Gaussian noise $\bm{\varepsilon}_n$ to their parameters. The noise is i.i.d. in each dimension of the parameter space, and it is scaled by the hyperparameter $\sigma$. The perturbed models are evaluated to obtain their reward scores $R_n$. The final update of the model parameters aggregates the sampled perturbations by weighting them using their normalized reward scores. The standard update equation $\bm{\theta}_{t} \leftarrow \bm{\theta}_{t-1}+ \alpha \cdot \frac{1}{\sigma}\frac{1}{N}\sum_{n=1}^N R_n \bm{\varepsilon}_n$ is simplified to $\bm{\theta}_{t} \leftarrow \bm{\theta}_{t-1}+ \alpha \cdot \frac{1}{N}\sum_{n=1}^N R_n \bm{\varepsilon}_n$ by digesting the term $\frac{1}{\sigma}$ into the learning rate $\alpha$.

To improve scalability, a number of modifications to this basic algorithm were made as detailed in the next section.
\subsection{Implementation details} \label{subsec:implementation}
\begin{table*}[t]
\centering
%\small
\begin{threeparttable}
\renewcommand{\arraystretch}{1}
\setlength{\tabcolsep}{5pt}

\rowcolors{3}{white}{gray!6}
\begin{tabular}{
  c   % Model
  c   % Baseline
  c   % PPO
  c   % GRPO pop8
  c   % GRPO pop30
  c   % GRPO-verl
  c   % Dr.GRPO-verl
  c   % ES
}
\toprule
\multirow{2}{*}{\textbf{Base Model}}
  & \multirow{2}{*}{\textbf{Original}}
  & \multicolumn{5}{c}{\textbf{RL}}
  & \multirow{2}{*}{\textbf{ES (ours)}} \\
\cmidrule(lr){3-7}
& & \textbf{PPO-z} & \textbf{$\text{GRPO-z (8)}$} & \textbf{$\text{GRPO-z (30)}$} &
\textbf{$\text{GRPO-v}$} & 
\textbf{$\text{Dr.GRPO-v}$} \\
\midrule
\rowcolor{green!3} Qwen-2.5-0.5B-Instruct & 0.1   & 0.3  & 0.3  & 0.5 & 13.0 & 13.5 & \textbf{14.4} \bfseries \\
\rowcolor{green!3} Qwen-2.5-1.5B-Instruct & 0.7   & 14.2  & 13.9 & 14.8 & 27.8 & 31.0 &  \textbf{37.3} \bfseries \\
\rowcolor{green!3} Qwen-2.5-3B-Instruct   &  10.0  & 20.1 & 30.9 & 32.5 & 37.8 & 43.8 & \textbf{60.5} \bfseries \\
\rowcolor{green!3} Qwen-2.5-7B-Instruct   & 31.2   & 55.1 & 54.2 & 52.8 & 57.0 & 57.5 & \textbf{66.8} \bfseries \\
\rowcolor{blue!3}  Llama-3.2-1B-Instruct  &  0.4  & 11.2  & 14.5  & 13.0 & 14.9 & 13.9 & \textbf{16.8} \bfseries \\
\rowcolor{blue!3}  Llama-3.2-3B-Instruct  & 3.2   & 35.3 & 39.4 & 38.8 & 42.5 & 47.8 & \textbf{51.6} \bfseries \\
\rowcolor{blue!3}  Llama-3.1-8B-Instruct  &  8.1  & 42.8  & 49.9  & 51.3  & 46.9 & 50.2 & \textbf{61.2} \bfseries \\
\bottomrule
\end{tabular}
\end{threeparttable}
\caption{Accuracy (\%) on the Countdown task across model families, sizes, and fine-tuning algorithms. Different model families are shaded for clarity; \textit{Original} refers to directly evaluating the base model without any fine-tuning, and GRPO-z (8) and GRPO-z (30) indicate group sizes of 8 and 30. The suffix "-z" and "-v" represents different implementation variants (see Appendix~\ref{app:exp_setup} for more details). The same hyperparameters were used for all ES runs; a separate grid search for the best hyperparameters was run for each RL experiment.
}
\label{tab:countdown_main}
\end{table*}
The actual implementation of ES for this paper expands on the basic ES algorithm in seven ways (see Algorithm~\ref{alg:es-llm_specific} for the detailed pseudocode):

(1) \emph{Noise retrieval with random seeds:} Similar to \citet{salimans2017es, such2017DeepNG}, only the random seeds are stored to reduce GPU memory usage. The perturbation noise used during sampling can be retrieved exactly by resetting the random number generator with specific random seeds. (2) \emph{Parallel evaluations:} In each iteration, the perturbed models can be evaluated fully in parallel by assigning a separate random seed to each process. (3) \emph{Layer-level in-place perturbation and restoration:} To reduce the peak GPU memory usage, the model parameters are perturbed in-place layer by layer, with corresponding random seeds archived. After evaluation of the perturbed model, the model parameters are restored by subtracting the same noise perturbations using the archived random seeds. For each evaluation process, apart from the model parameters, the only additional memory needed is to store a tensor the size of a layer temporarily. (4) \emph{Reward normalization:} The rewards of the perturbed models are normalized using $z$-score within each iteration, so that the normalized rewards for each iteration have a mean of 0 and standard deviation of 1. This normalization makes the reward scale consistent across iterations and tasks. (5) \emph{Greedy decoding:} The perturbed models use greedy decoding to generate the responses for reward evaluations. As a result, the perturbed models are evaluated deterministically, so that all performance differences come from the exploration in parameter space instead of action space. (6) \emph{Decomposition of the parameter update:} At the end of each iteration, the aggregated update of model parameters is performed in-place in a decomposed manner, gradually adding up layer by layer and seed by seed, significantly reducing the peak GPU memory needed. (7) \emph{Learning rate digestion:} The standard update equation $\bm{\theta}_{t} \leftarrow \bm{\theta}_{t-1}+ \alpha \cdot \frac{1}{\sigma}\frac{1}{N}\sum_{n=1}^N R_n \bm{\varepsilon}_n$ is simplified to $\bm{\theta}_{t} \leftarrow \bm{\theta}_{t-1}+ \alpha \cdot \frac{1}{N}\sum_{n=1}^N R_n \bm{\varepsilon}_n$ by digesting the term $\frac{1}{\sigma}$ into the learning rate $\alpha$, simplifying the computation and parametric setup.

To highlight the strength of ES, %\sout{In order to keep the algorithm simple}
we intentionally remove common algorithmic enhancements explored in OpenAI ES \citep{salimans2017es}. Enhancements like %\sout{such as}
rank transformation of rewards \citep{wierstra14a}, mirrored sampling \citep{sehnke2010parameter}, weight decay, and virtual batch normalization \citep{salimans2016gan} are not used in this work. Additionally, we do not utilize %\sout{and neither are} 
more advanced optimizers like Adam \citep{kingma2015adam}. %\sout{They can be included in to improve results in future work.} 
This design choice isolates the core ES algorithm and demonstrates that strong performance can be achieved without auxiliary enhancements. In future work, each individual enhancement can be explored to further improve performance.
\section{Empirical Studies}
This section first compares the fine-tuning performance of ES and RL baselines on a standard reasoning benchmark. After that, behavioral differences between ES and RL are investigated in fine-tuning for conciseness, followed by comparisons to more SOTA RL baselines on several math reasoning tasks. Finally, ES is applied to solve two challenging puzzle problems.

\label{sec:exp}
\subsection{Performance in the Countdown task}
Fine-tuning performance was measured in the Countdown task~\citep{gandhi2024stream, tinyzero}, a symbolic reasoning benchmark (see Appendix~\ref{app:countdown_example} for details), showing that ES is accurate and efficient across different kinds and sizes of LLMs, even when the RL approaches are not.

\textbf{Experimental setup.}\label{sec4:addi_imple}
A single fixed set of hyperparameters ($N=30$, $\sigma=0.001$, $\alpha=5 \times 10^{-4}$) was used for all ES Countdown experiments. Notably, the population size 30 is significantly lower than those in previous works \citep{salimans2017es,zhang2017essgd}, in which $N\geq10,000$. For RL baselines (see Appendix~\ref{app:exp_setup} for details), a separate hyperparameter sweep was done for each experiment. RL methods turned out sensitive to hyperparameters, in particular the KL-divergence penalty coefficient $\beta$ and learning rate $\alpha$, and did not make much progress if they were not set precisely. To mitigate this issue, for each model, a small grid of $\beta$ and $\alpha$ values were tested and the best-performing configuration selected (see Table~\ref{tab:ppo_grpo_sweep} in the Appendix~\ref{app:exp_setup}). This approach makes the comparison conservative with respect to ES, but it also highlights its robustness. 
\begin{table*}[t!]
\centering
%\small
\renewcommand{\arraystretch}{1}
\setlength{\tabcolsep}{6pt}

\rowcolors{3}{white}{gray!6}
\begin{tabular}{
  l
  c
  c
  c
  c
  c
}
\toprule
\textbf{Model} & \boldsymbol{$\beta$} & \boldsymbol{$\alpha$} & \boldsymbol{$\sigma$} & \textbf{Reward} $\uparrow$ & \textbf{KL} $\downarrow$ \\
\midrule
%\rowcolor{red!3} Qwen-2.5-7B & \xmark  & \xmark & \xmark & $1.0\pm0.01$   & $1.0\pm0.01$  \\
\rowcolor{green!3} Qwen-2.5-7B+GRPO & 0.0  & $5\times10^{-6}$  & \xmark & $0.867\pm0.054^{*}$   & $0.861\pm0.614^{*}$  \\
\rowcolor{green!3} Qwen-2.5-7B+GRPO & 0.01  & $5\times10^{-6}$  & \xmark & $0.871\pm0.060^{*}$   & $1.354\pm0.873^{*}$  \\
\rowcolor{green!3} Qwen-2.5-7B+GRPO & 0.0167  & $5\times10^{-6}$  & \xmark & $0.911\pm0.038$   & $1.591\pm0.811$  \\
\rowcolor{green!3} Qwen-2.5-7B+GRPO & 0.0464  & $5\times10^{-6}$  & \xmark & $0.881\pm0.062$   & $1.384\pm1.187$  \\
\rowcolor{blue!3} Qwen-2.5-7B+ES & \xmark  & $0.0005$  & $0.001$ & $0.889\pm\textbf{0.004}$   & $0.274\pm\textbf{0.096}$  \\
\rowcolor{blue!3} Qwen-2.5-7B+ES & \xmark  & $0.00075$  & $0.0015$ & $0.919\pm\textbf{0.008}$   & $0.813\pm\textbf{0.212}$  \\
\bottomrule
\end{tabular}
\caption{Behavior or GRPO and ES in terms of mean conciseness reward and mean KL divergence. The label $^{*}$ indicates cases where reward hacking was observed. Only models that did not hack the reward were included in the results.
\label{table:es_grpo_mean_std}
}\end{table*}

\textbf{ES improves upon RL baselines across all tested models.} Previously, \citet{gandhi2025cognitive} found that RL does not generalize well across models on the Countdown task. Table~\ref{tab:countdown_main} confirms this result, and also demonstrates that ES does not have this problem.  With each model in the Qwen2.5 family (0.5B–7B) and the Llama3 family (1B–8B), ES substantially improved over PPO, GRPO and Dr.GRPO \cite{liu2025understanding}, including their implementation variants, often by a large margin (see Figure~\ref{fig:acc_improv_over_base} in Appendix~\ref{appendix:train_curve} for a model-wise visual comparison). Additional experiments on performance variances across multiple runs are provided in Appendix~\ref{app:additional_countdown}, showing a clear performance separability between ES and RL. These results demonstrate that ES scales effectively across different model types and sizes, and does so significantly better than RL.

\subsection{Behavioral differences between ES and RL in fine-tuning for conciseness}
\label{sec:behavioral_differences}
In order to characterize the different approaches that ES and RL take, they were used to fine-tune Qwen-2.5-7B Instruct, towards more concise responses in question-answering (see Appendix~\ref{app:exp_setup} for more details). That is, fine-tuning was rewarded based on how concise the answers were, but not directly rewarded for its question-answering performance. In this setup, it was possible to analyze not only whether fine-tuning was effective, but also how it was achieved, including what its side effects were.

\begin{figure}
\includegraphics[width=\linewidth]{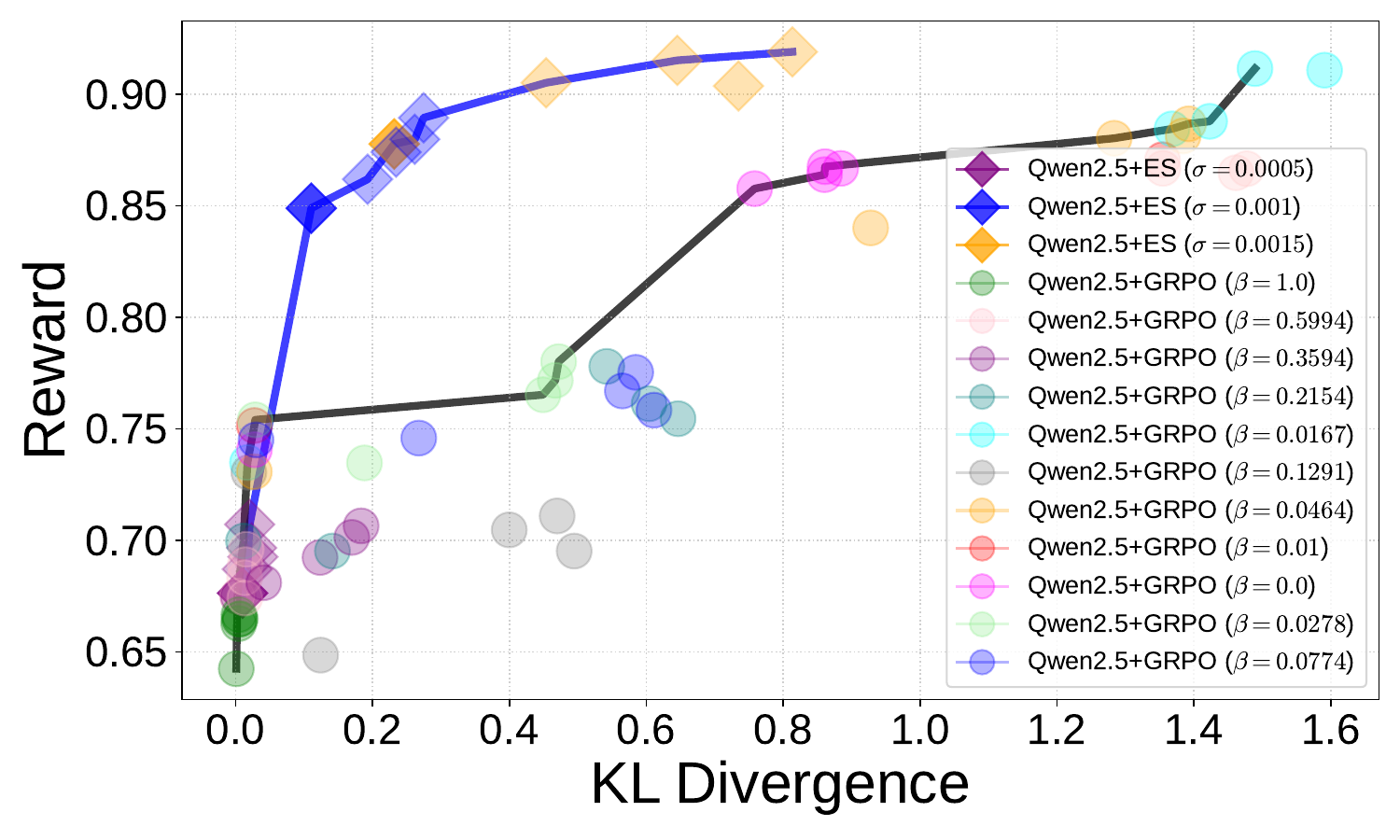}
\caption{Mean conciseness reward and mean KL divergence from the base model for each fine-tuning checkpoint across different learning parameters. The Pareto front of ES (blue line) is higher and to the left of the GRPO Pareto front (black line) models, indicating that it found better tradeoffs. ES discovers these solutions without any KL divergence penalty, suggesting that it represents a distinctly different fine-tuning mechanism from the GRPO.}
\label{fig:es_grpo_pareto_front}
\end{figure}

\textbf{ES discovers a dominant Pareto front.} Similarly to \citet{rafailov2023dpo}, a Pareto frontier analysis was used to compare ES and GRPO, with mean reward and mean KL divergence as the metrics (Figure \ref{fig:es_grpo_pareto_front}). The experimental setup is described in Appendix~\ref{app:exp_setup}.
% We highlight the Pareto optimal ES (blue line) and GRPO (black line) models with respect to the mean reward and KL divergence using $4$ random seeds in Figure \ref{fig:es_grpo_pareto_front}.
The ES Pareto front is represented by a blue line on top and the GRPO Pareto front by the black line below. That is, ES produced better tradeoffs than GRPO, i.e.\ models with higher reward and lower KL divergence. The GRPO results were achieved only after augmenting the conciseness reward with a KL divergence penalty (weighted by a parameter $\beta$). Without it, fine-tuning resulted in excessive divergence and incorrect answers. Remarkably, ES achieved superior tradeoffs without any KL divergence penalty, suggesting that ES fine-tuning is based on discovering distinctly different kinds of solutions than GRPO. Appendix \ref{appendix:extended_conciseness_exp} presents additional experiments with varying $\alpha$ and $\beta$ values, yielding similar conclusions.

\textbf{ES is more robust against reward hacking.} GRPO with $\beta=\{0.0, 0.01\}$ sometimes hacked the reward. Reward hacking was detected quantitatively using  the KL divergence estimator defined in \citet{schulman2020approximating}. A model was automatically marked as a “possible anomaly” if the KL divergence exploded by at least one order of magnitude. For these “possible anomaly” models, response details were checked to confirm the reward hacking behavior. Models which hacked the reward typically produced responses that were short but contain nonsensical symbols rather than words. By increasing the KL-penalty via higher $\beta$ values, reward hacking could be prevented. The optimal $\beta$ is likely to be problem specific and to require extensive search to find. In contrast, ES does not receive any feedback about the divergence of the fine-tuned model, and only seeks to optimize conciseness. Regardless, it did not exhibit any reward hacking, despite achieving mean reward comparable to GRPO with $\beta=\{0.0, 0.01\}$. This result again suggests that ES finds a different way of optimizing the reward function.
\begin{figure*}
\centering
\includegraphics[width=\linewidth]{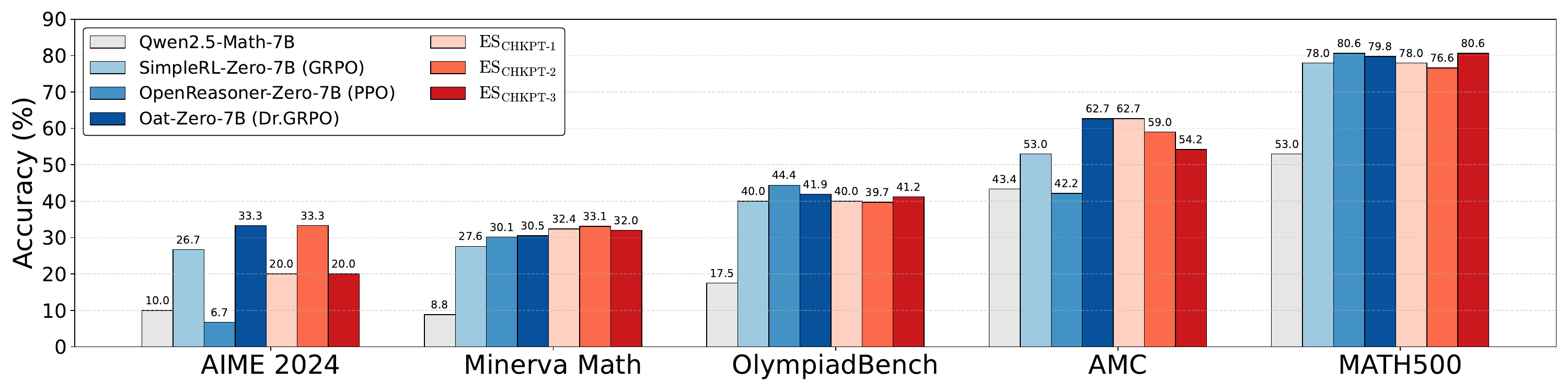}
\caption{Performance of ES compared to strong, well-established RL baselines across math reasoning benchmarks. Across all benchmarks, ES achieved competitive performance compared to OpenReasoner-Zero-7B (PPO), Simple-RL-Zero (GRPO), Oat-Zero-7B (Dr.GRPO). Given the vanilla nature of the current ES implementation, these results constitute a promising starting point for ES fine tuning in math.}
\label{fig:math_experiments}
\end{figure*}

\textbf{ES fine-tuning is reliable across runs.} Fine-tuning LLMs is computationally expensive, so it is critical that it leads to consistent results across runs. Table \ref{table:es_grpo_mean_std} presents the mean and standard deviation of the conciseness reward and KL divergence across four independent runs after $1,000$ iterations. A mean reward cut-off of $>0.85$ was used to down-select hyperparameter combinations, ensuring that only the best ES and GRPO configurations were included in the analysis. From Table~\ref{table:es_grpo_mean_std}, ES achieved consistent conciseness rewards, indicated by a low standard deviation over four runs with different random seeds. GRPO has $15.5\times$ higher standard deviation, suggesting that its results were much less consistent. The results on KL divergence show similar patterns. Thus, ES fine-tuning is more reliable than GRPO. %For instance, while ES ($\sigma=0.0015$) and GRPO ($\beta=0.0167$) achieved similar mean rewards, GRPO exhibits a $1.95\times$ higher KL divergence mean and $3.83\times$ greater standard deviation. Similarly, while ES ($\sigma=0.001$) achieves a slightly lower reward compared to GRPO ($\beta=0.0167$), GRPO ($\beta=0.0167$) has a $5.8\times$ higher KL divergence and a $8.44\times$ higher standard deviation. Thus, ES fine-tuning is more reliable than GRPO.

\subsection{ES applied to Math reasoning tasks}
RL has been shown to enhance the reasoning capabilities of LLMs through post-training with verifiable rule-based rewards. To understand the impact of ES on LLM reasoning, ES fine-tuning was evaluated on a set of standard math benchmarks from the literature. The main result is that ES is competitive with SOTA RL in this setting.

\textbf{Training setup.} The Qwen2.5-Math-7B \cite{yang2024qwen2} base model was fine-tuned with ES using the MATH dataset \cite{hendrycks2021measuring}. Problems labeled with difficulty ranging from 3-5 were included, and the Qwen Math template was used for training with ES \citep[see Appendix \ref{app:math_extended};][]{yang2024qwen2}. Both RL and ES sampled a maximum of $3,000$ tokens per response; ES hyperparameters were set to $\sigma=0.001$, $\alpha=\frac{\sigma}{2}$, and $N=30$.
%\textbf{Training setup.} We fine-tune the Qwen2.5-Math-7B \cite{yang2024qwen2} base model with ES using the MATH dataset \cite{hendrycks2021measuring}. Specifically, we include problems labeled with difficulty ranging from 3-5. Additionally, we use the Qwen Math template for training with ES. For ES we set $\sigma=0.001$, $\alpha=\frac{\sigma}{2}$, and $N=30$. We allow ES to sample a maximum of $3,000$ tokens per response.
%\textbf{Training setup.} The experiments utilized the Qwen2.5-Math-7B base model with the Qwen Math template (see Appendix \ref{app:math_extended}) \cite{yang2024qwen2}. They focused on problems with difficulty 3 to 5 in the MATH dataset \cite{hendrycks2021measuring} as described below. Both RL and ES sampled a maximum of $3,000$ tokens per response; ES hyperparameters were set to $\sigma=0.001$, $\alpha=\frac{\sigma}{2}$, and $N=30$.

\textbf{RL baselines.}
The fine-tuned ES models were compared with strong, well-established baselines from the literature. These RL implementations achieve the most SOTA performance in the tested benchmarks, utilizing production-ready RL libraries like VERL \cite{sheng2024hybridflow} and OAT \cite{liu2024oat}. They include SimpleRL-Zero (GRPO) \cite{zeng2025simplerl}, OatZero (Dr.GRPO) \cite{liu2025understanding}, and OpenReasoner (PPO) \cite{hu2025open}. The publicly released Qwen2.5-7B checkpoints trained with the original training recipes were used for evaluation. Note that SimpleRL-Zero and OatZero were trained using the MATH dataset \cite{hendrycks2021measuring}, whereas OpenReasoner was trained using a custom dataset compiled by the authors. Consequently, performance differences should be interpreted in light of both algorithmic and dataset-related differences.

\textbf{Evaluation benchmarks.}
Several standard math reasoning benchmarks from the literature are used for evaluation: OlympiadBench \cite{he2024olympiadbench}, MATH500 \cite{hendrycks2021measuring}, Minerva \cite{lewkowycz2022solving}, AIME2024 \cite{li2024numinamath}, and AMC \cite{li2024numinamath}. The pass@1 accuracy metric was used in the evaluations.

\textbf{Key results.} Figure~\ref{fig:math_experiments} shows the performance of the base model, three checkpoints of RL baselines, and three checkpoints of ES (see Appendix~\ref{app:math_extended} for more details). ES significantly improved the base models across each benchmark, showing the other optimization methods aside from RL can be used to elicit improvement in LLM reasoning capabilities. In addition, ES exhibits competitive performances compared with the SOTA RL baselines in all the benchmarks. It is notable that these RL baselines are the best-performing implementations selected from the literature, with extensive algorithmic refinement and hyperparameter search particularly for math reasoning tasks. In contrast, the current ES implementation is a vanilla variant with a simple hyperparameter setup; thus, the results constitute a promising starting point for the ES approach in math fine tuning.

\subsection{Solving challenging puzzle problems} \label{subsec:puzzle}

To further evaluate the generality of ES in tackling different types of tasks, two challenging puzzle problems were used as additional testbeds. The first is ARC-AGI~\citep{chollet2024arc}, a benchmark designed to evaluate fluid intelligence \cite{chollet2019measureintelligence}. The second is Sudoku, a logic-based number-placement puzzle. Whereas the base LLM models fail severely in both problems, ES fine-tuning significantly improves their performance (Table~\ref{tab:reasoning_main}). Experimental details are provided in Appendix~\ref{app:arc-agi} for ARC-AGI and Appendix~\ref{app:sudoku} for Sudoku.

\begin{table}[h]
\centering
%\small
\begin{threeparttable}
\renewcommand{\arraystretch}{1}
\setlength{\tabcolsep}{5pt}
\begin{tabular}{llcc}
\toprule
\textbf{Task} & \textbf{Base Model} & \textbf{Original} & \textbf{ES Fine-tuned} \\
\midrule
ARC-AGI & Qwen-2.5-14B & 0.2 & \textbf{29.5} \\
Sudoku  & Qwen-2.5-3B  & 2.5 & \textbf{69.5} \\
\bottomrule
\end{tabular}
\end{threeparttable}
\caption{Accuracy (\%) on solving puzzle problems. \textit{Original} refers to directly evaluating the base model without any fine-tuning. ES fine-tuning improves performance significantly, demonstrating that it can be applied to a range of problems.}
\label{tab:reasoning_main}
\end{table}

\section{Discussion} \label{sec:discussion}
\textbf{Algorithmic advantage of ES vs.\ RL.} Exploration in parameter space plays a key role in the surprisingly good fine-tuning performance of ES. As discussed by \citet{thomas2010parameter} and \citet{plappert2018parameter}, sampling noise in parameter space ensures that the entire action trajectory, i.e., the sequence of tokens, only depends on one single sampling, leading to significantly lower variance in rollouts, i.e., in response generation. As a result, gradient estimation is more reliable and convergence is more stable. In contrast, action space exploration in RL injects noise at every step, i.e., at each token position, resulting in high variance in the sequence generation. The behavior of RL therefore is much less reliable than ES, as was seen in Table~\ref{table:es_grpo_mean_std}. Moreover, step-wise exploration in action space promotes reward hacking by increasing the chance of sampling a single hacking action. One example is the nonsensical symbol sampled during RL that can hack the conciseness reward.

Another key difference between ES and RL is that ES intrinsically optimizes a solution distribution \citep{lehman2018esmore}, while RL optimizes a single solution. This property makes it more difficult for ES to hack the reward since a single hacked solution usually does not have a high-quality solution distribution around it. This property also results in solutions that are more robust to noisy perturbations in parameter space \citep{lehman2018esmore}, making them more robust to adversarial attacks and less likely to be compromised in other follow-up fine-tuning tasks \citep{chen2025loss}.

%In the experiments in this paper, extensive hyperparameter tuning was performed for RL methods, resulting in specific RL hyperparameters for different model sizes and families. In comparison, ES was found to be less sensitive to hyperparameters, and the same set of hyperparameters was used for all experiments. While there are many common enhancements for ES \citep{salimans2017es}, none were used in the experiments so that the power of vanilla ES could be clearly demonstrated. Thus, it may be possible to improve the results with more extensive hyperparameter tuning and other enhancements.

The ES algorithm presented in Algorithm \ref{alg:es-llm_specific} is very simple and easy to implement, without need for sophisticated hyperparameter search. In the current experiments, ES with a single fixed set of hyperparameters was able to achieve robust performances across different tasks and base models. Ablation studies in Appendix~\ref{app:hyperparam_ablation} further demonstrate the insensitivity of ES to hyperparameters. In contrast, RL algorithms are considerably more complex and require substantial expertise to implement robustly across tasks and systems, usually with extensive hyperparameter tuning. In particular, there has been significant debate in the literature regarding best practices for implementing GRPO. Effective GRPO implementations typically rely on a number of non-obvious design choices and implementation details, such as removing length normalization \cite{liu2025understanding} and using more aggressive clipping \cite{yu2025dapo}. Many of these practices have only emerged through extensive empirical investigation. Moreover, the application of the KL penalty in GRPO remains an open design choice, with alternatives such as applying it to the loss or directly to the reward leading to markedly different performance outcomes \cite{shah2025comedy}

\textbf{Engineering benefits of ES vs.\ RL.}
Modern RL frameworks grow increasingly complex as they are applied to LLMs with ever-increasing parameter counts. Deploying these systems in practice often requires substantial engineering effort and computational resources. In contrast, ES is simple to implement and can help democratize post-training by significantly lowering engineering and systems overhead. This section outlines two key advantages of ES and suggests how it can be scaled to fine-tune the largest LLMs.

(1) \emph{Parallelization.} To minimize memory overhead and maximize sample throughput, RL systems rely on asynchronous architectures in which actors are distributed across GPUs and update a shared learner model. While effective, scaling these systems across large numbers of GPUs and computational nodes introduces significant engineering complexity. In contrast, as shown by \citet{salimans2017es}, ES can be trivially parallelized: as the number of available GPUs increases, the population size can be scaled accordingly.

Modern frontier AI research labs operate clusters with thousands of GPUs\footnote{LLama3 training used $16,000$ H100 GPUs \cite{llama3modelcard}.}, making efficient large-scale parallelization possible. While it is challenging to do with RL, ES requires only the exchange of random seeds (noise) and scalar rewards between machines. Such simple communication enables parallel ES to be used either to reduce wall-clock training time or to scale to much larger populations.

(2) \emph{Gradient computation.} Asynchronous RL makes it possible to compute actor-related gradients in parallel. However, in order to manage memory usage, gradient checkpointing and multiple learner updates per synchronization step are needed. While these techniques enable larger effective batch sizes, they also require gradients to be communicated across GPUs, and sometimes nodes, introducing significant memory overhead and engineering complexity. This complexity scales with both the number of GPUs and the size of the model, and is further exacerbated when model parameters must be sharded across devices.

In contrast, ES does not require gradient computation. By eliminating gradient calculation and communication entirely, ES avoids much of the associated engineering and memory overhead. As a result, each member of the ES population can use large batch sizes freely without cross-device gradient synchronization, which can yield substantial practical and performance benefits. Moreover, the computational cost in terms of FLOPs is significantly lower for ES compared to RL methods, which involve backpropagation and additional reference models (See Appendx~\ref{app:computation} for detailed analysis).

Importantly, ES is an inference-only fine-tuning mechanism, where the model weights are never differentiated, only evaluated. This property opens the door to specialized inference kernels optimized for repeated forward passes, large batches, and parameter perturbations. These mechanisms are difficult to leverage in gradient-based training regimes, but are possible in ES fine tuning in the future.

%\emph{(3) RL algorithmic implementation.} The ES algorithm presented in Algorithm \ref{alg:es-llm_specific} is very simple and easy to implement. In contrast, RL algorithms are considerably more complex and require substantial expertise to implement robustly across tasks and systems. In particular, there has been significant debate in the literature regarding best practices for implementing GRPO. Effective GRPO implementations typically rely on a number of non-obvious design choices and implementation details, such as removing length normalization \cite{liu2025understanding} and using more aggressive clipping \cite{yu2025dapo}. Many of these practices have only emerged through extensive empirical investigation. Moreover, the application of the KL penalty in GRPO remains an open design choice, with alternatives such as applying it to the loss or directly to the reward leading to markedly different performance outcomes \cite{shah2025comedy}.

\vspace*{1ex}
\section{Future Work} \label{sec:futurework}

One counterintuitive result is that the ES implementation only needs a population of 30 to effectively optimize billions of parameters. In contrast, previous work \citep{salimans2017es,zhang2017essgd,lehman2018esmore,lorenc2025estransformer} used populations of 10,000 or more for models with millions or fewer parameters. An interesting future direction is to analyze how such small populations are possible. Perhaps this is related to the observed low intrinsic dimensionality of LLMs \citep{aghajanyan2021intrinsic}. Another promising direction is to use ES to perform unsupervised fine-tuning based on internal behaviors of LLMs, such as confidence calculated based on semantic entropy and semantic density \citep{qiu:neurips24,farquhar:nature24}. Such fine-tuning cannot be done with RL, since action space exploration does not change the internal representations of LLMs (that is, each action sampling is generated via output distribution without changing the internal parameters). In a broader sense, since ES does not need process rewards during exploration, it may be a necessary ingredient for superintelligence \citep{mucci2023super}, which would be difficult to achieve by supervised learning using process guidance from human data. Massive parallelization of ES will speed up exploration by distributing the computations across GPU machines or even data centers.

% jaggedness discussion
An important question is: what are the underlying computational mechanisms that make ES and RL behave so differently? While this question requires significant further work, a possible hypothesis emerges from the experiments in this paper. Many fine-tuning objectives, like conciseness and the Countdown task, are long-horizon outcome-only objectives. The reward signal is jagged, making it difficult to navigate with gradient-based post-training methods. RL and ES both provide workarounds via effective noise injection to ``smooth out’’ the jagged reward landscape. In the case of RL, noise is introduced from Monte-Carlo sampling of each token during a rollout, averaged over many rollouts, which effectively smooths the sampling process but does not necessarily guarantee that the reward landscape is smooth in parameter space. RL's gradient estimation therefore has a high-variance, %\conor{It might be good to include a citation here for high-variance on the RL gradient?}
and its signal-to-noise ratio becomes worse with longer sequences and sharper policies (i.e.\ those with lower entropy), and therefore prone to undesirable outcomes such as reward hacking. %\conor{Are there any papers on the causes of reward hacking we can cite?}

In contrast, ES injects noise directly into the parameter space via explicit Gaussian convolution, which effectively smooths out the jagged reward landscape. %\conor{citation?}
As a result, it provides a more stable way of exploring the landscape, leading to more consistent, efficient, and robust optimization (as observed in the experiments and in Appendix~\ref{app:shifts}). Moreover, the larger the models and the sharper the policies, %\conor{What is a sharp policy?}
the more jagged the reward landscapes; therefore, ES is likely to have an advantage in fine-tuning them. Direct evidence for this hypothesis still needs to be obtained, but it provides a plausible mechanistic explanation, and a direction for future work. 

In principle, ES and RL are not mutually exclusive. They can be complementary to each other and used in a combined way. One possibility is to divide the learning process into exploration and exploitation phases, and then alternate applying ES and RL. For global exploration, perturbation-based parameter space search is more explorative and less likely to get stuck in an initial local optimum. For fine-grained local search, following the gradients is a more accurate and effective approach. Investigating synergies between ES and RL could result in better fine-tuning methods, as well as an improved understanding of LLM training in general.

\section{Conclusion}

This paper introduces a fundamentally new paradigm for fine-tuning LLMs by scaling ES to models with billions of  parameters without dimensionality reduction. Contrary to long-standing assumption that such scaling is infeasible, the paper demonstrates that ES can efficiently fine-tune the full parameter space of modern LLMs and, in doing so, consistently surpasses standard RL-based fine-tuning methods. On the Countdown task, with sparse long-horizon rewards challenging for gradient-based RL, ES achieves substantially stronger performance. It also exhibits markedly reduced sensitivity to hyperparameter choices and delivers stable, repeatable improvements across multiple base LLMs. In fine-tuning for conciseness, ES is less prone to reward hacking and shows reliable behavior across independent runs. The generality of ES fine tuning is further validated by strong performance on state-of-the-art math reasoning benchmarks and two challenging puzzle problems. Together, these results establish ES as a scalable, robust, and general fine-tuning method, and demonstrate that backpropagation-free optimization can serve as a powerful alternative to RL for fine-tuning LLMs.

%\conor{Again, see my comment in the overleaf comments section} This paper introduces and evaluates a new approach to fine-tuning LLMs, based on scaling up ES to billions of dimensions. The approach performed significantly better than the standard RL fine-tuning in the Countdown task, which has sparse long-horizon rewards. ES was found to be less sensitive to hyperparameter setup, and to achieve consistently better results across multiple LLMs. Empirical studies on fine-tuning for conciseness revealed that ES is less likely to hack the reward, and behaves reliably across multiple runs. The generalizability of ES is further verified by its strong performances in several SOTA math reasoning tasks and two challenging puzzle problems. %The mechanisms underlying these differences still need to be characterized, but a plausible hypothesis is that the exploration in ES is better suited for the jagged reward landscapes in large models. ES therefore constitutes a promising alternative to RL in fine-tuning LLMs.

\section*{Impact Statement}

Beyond the standard potential consequences of advancing the field of machine learning, there are two key areas of broader impact, stemming from (1) \emph{increased ease of use} and (2) \emph{reduced reward-hacking}.

\textbf{Ease of use:}
ES qualitatively reduces the barrier of entry to fine-tuning LLMs.
Unlike RL, which requires an expert mathematical understanding of nuanced gradient-based training dynamics to design an effective reward function, ES simply requires the experimenter to assign \emph{a score} to a model after it has attempted a task.
This simplification democratizes LLM fine-tuning, opening the door to the development of customized AI applications by non-experts.

\textbf{Reward-hacking:}
As shown in Section~\ref{sec:behavioral_differences} and prior work \cite{lehman2018esmore}, ES is inherently less susceptible to reward-hacking than RL and other gradient-based methods.
Thus, LLMs fine-tuned with ES are less likely to lose ethical guardrails present in the base model. Similarly, it may be easier to fine-tune for ethical behavior (i.e.\ alignment) with ES, since the model is less likely to overfit to specific training examples.

Combining the above two areas of impact, ES fine-tuning can reduce the risk of unintended ethical misbehavior of LLMs fine-tuned by non-experts.

\section*{Acknowledgments}
We would like to thank Sid Stuart for providing technical support for hardware management and always being responsive. We would like to thank Kajetan Schweighofer for his generous help during the peer-review rebuttal phase and insightful discussions. We would like to thank Jamieson Warner for providing valuable feedback.
% In the unusual situation where you want a paper to appear in the
% references without citing it in the main text, use \nocite
%\nocite{langley00}

\bibliography{example_paper}
\bibliographystyle{icml2026}

%%%%%%%%%%%%%%%%%%%%%%%%%%%%%%%%%%%%%%%%%%%%%%%%%%%%%%%%%%%%%%%%%%%%%%%%%%%%%%%
%%%%%%%%%%%%%%%%%%%%%%%%%%%%%%%%%%%%%%%%%%%%%%%%%%%%%%%%%%%%%%%%%%%%%%%%%%%%%%%
% APPENDIX
%%%%%%%%%%%%%%%%%%%%%%%%%%%%%%%%%%%%%%%%%%%%%%%%%%%%%%%%%%%%%%%%%%%%%%%%%%%%%%%
%%%%%%%%%%%%%%%%%%%%%%%%%%%%%%%%%%%%%%%%%%%%%%%%%%%%%%%%%%%%%%%%%%%%%%%%%%%%%%%
\newpage
\appendix
\onecolumn
\section{Appendix}
\subsection{Experimental Setup} \label{app:exp_setup}

\paragraph{Experimental setup for the Countdown experiments.} Representative models from the Qwen2.5 family (0.5B–7B) and the Llama3 family (1B–8B) were fine-tuned for this task. For the PPO-z experiments, a grid search was first performed around common hyperparameter settings and the best-performing values used (Table~\ref{tab:ppo_grpo_sweep}). TinyZero (\href{https://github.com/Jiayi-Pan/TinyZero}{https://github.com/Jiayi-Pan/TinyZero}) is used for PPO-z implementations. For the GRPO-z experiments, a grid search was performed around the settings of \cite{tinyzero} and the best-performing values used. GRPO-z experiments were run with two different group sizes: $N=8$, following the common practice in GRPO training for the Countdown task, and $N=30$, aligning with the population size in ES. GRPO-Zero (\href{https://github.com/policy-gradient/GRPO-Zero}{https://github.com/policy-gradient/GRPO-Zero}) is used for GRPO-z implementations. VERL \cite{sheng2024hybridflow} is used for both GRPO-v and Dr.GRPO-v implementations, with the standard default configurations for math reasoning benchmarks. 

% added
For the VERL implementations, we set the global batch size of 1024, a learning rate of $1 \times 10^{-6}$, and a rollout group size of $N=8$. We compared two configurations: GRPO-v and Dr.GRPO-v. The GRPO-v baseline incorporated a standard KL divergence penalty with a coefficient of $\beta=0.001$. In contrast, the Dr.GRPO-v configuration removed the KL penalty (\texttt{use\_kl\_loss=False}) and disabled advantage normalization (\texttt{norm\_adv\_by\_std=False}). Instead, Dr.GRPO-v employed a sequence-mean token-sum normalization strategy for loss aggregation with a scaling factor of 1024.
%--

For all the ES and RL baselines, the total number of sample evaluations was the same. The ES population size was $N=30$, noise scale $\sigma=0.001$, and learning rate $\alpha=5 \times 10^{-4}$ across all experiments.  
To evaluate accuracy, a set of 200 samples were used during training, and a different set of 2000 samples during testing. For ES, results were reported on the test set after training for 500 iterations. For RL, the training was stopped after the same total number of sample evaluations as in the ES runs. An example of the prompt and the response is provided in Appendix~\ref{app:countdown_example}.

\begin{table}[h]
\centering
\setlength{\tabcolsep}{4pt}
\begin{tabular}{llcccc}
\toprule
Method & Model & ($1\mathrm{e}{-3}$, $1\mathrm{e}{-6}$) & ($1\mathrm{e}{-3}$, $1\mathrm{e}{-5}$) & ($5\mathrm{e}{-3}$, $1\mathrm{e}{-6}$) & ($5\mathrm{e}{-3}$, $1\mathrm{e}{-5}$) \\
\midrule
\multirow{7}{*}{PPO-z} 
& Qwen-0.5B-Instruct & \textbf{\checkmark} &  &  &  \\
& Qwen-1.5B-Instruct & \checkmark &  &  &  \\
& Qwen-3B-Instruct   & \checkmark &  &  &  \\
& Qwen-7B-Instruct   &  & \checkmark &  &  \\
& Llama-1B-Instruct  &  & \checkmark &  &  \\
& Llama-3B-Instruct  &  &  &  & \checkmark \\
& Llama-8B-Instruct  &  &  & \checkmark &  \\
\midrule
\multirow{7}{*}{GRPO-z} 
& Qwen-0.5B-Instruct &  & \checkmark &  &  \\
& Qwen-1.5B-Instruct &  &  & \checkmark &  \\
& Qwen-3B-Instruct   &  & \checkmark &  &  \\
& Qwen-7B-Instruct   & \checkmark &  &  &  \\
& Llama-1B-Instruct  &  &  &  & \checkmark \\
& Llama-3B-Instruct  &  & \checkmark &  &  \\
& Llama-8B-Instruct  & \checkmark &  &  &  \\
\bottomrule
\end{tabular}
\caption{Hyperparameter Sweep across Models under PPO-z and GRPO-z. 
Each pair $(\cdot,\cdot)$ denotes (KL-divergence penalty coefficient $\beta$, learning rate $\alpha$); 
the label '$\checkmark$' indicates the best hyperparameter setting for each model-method combination.}
\label{tab:ppo_grpo_sweep}
\end{table}

\paragraph{Experimental setup for the conciseness experiments.} In each experiment, Qwen-2.5-7B-Instruct~\citep{yang2025qwen2} was fine-tuned using both ES and GRPO and evaluated using a held-out evaluation set. Each run was repeated four times, using a different random seed each time.
For each GRPO experiment, the group size $N=30$, and learning rate $\alpha=5\times10^{-6}$. Ten log-spaced values from $0.01$ to $1.0$ were evaluated for the the KL-divergence penalty coefficient $\beta$, as well as $\beta=0.0$. Appendix \ref{appendix:extended_conciseness_exp} presents additional experiments with varying $\alpha$ and $\beta$ values. 
For ES, the population size $N = 30$, ensuring that GRPO and ES generated the same number of responses per prompt, resulting in the same training exposure. Models were fine-tuned with $\sigma = \{0.0005, 0.001, 0.0015\}$, with a learning rate $\alpha = \frac{\sigma}{2}$.
Both GRPO and ES experiments were run for $1,000$ iterations, and a checkpoint saved every $200$ iterations. Table \ref{table:conciseness_training_dataset} shows the dataset of prompts and verifiable solutions used during fine-tuning; note that it consists of only two examples. Similarly, Table \ref{table:conciseness_eval_dataset} lists the prompts and verifiable solutions used in evaluating each fine-tuned model. For all the experimental results, the displayed reward values are normalized to be within $[0,1]$, with $0$ corresponding to $-2000$ in the original reward function and $1$ corresponding to the best possible original reward $0$.

\begin{table}[t]
    \centering
    
    \begin{tabular}{|c|c|}
         \hline
         Prompt &  Verifiable Solution \\
         \hline
         Solve: $3 + 5 =$ & $8$ \\
         If all birds can fly and penguins are birds, can penguins fly & No \\
         \hline
    \end{tabular}
    \caption{Prompts and verifiable solutions used in fine-tuning the models for conciseness. Two examples is enough to achieve this goal.}
    \label{table:conciseness_training_dataset}
\end{table}

\begin{table}[]
    \centering
    
    \begin{tabular}{|c|c|}
         \hline
         Prompt &  Verifiable Solution \\
         \hline
         What is the capital of France? & Paris \\
         
         Calculate: $12 \times 7 =$ & $84$ \\
         
         Is the statement ``All cats are mammals'' true or false? & True \\
         What comes next in the sequence: $2, 4, 6, 8,$ ? & $10$ \\
         Translate ``Hello'' to Spanish: & Hola \\
         What is $15\%$ of $200$? & $30$ \\
         Name one primary color: & Red \\
         How many days are in a week? & $7$ \\
         \hline
    \end{tabular}
    \caption{Prompts and verifiable solutions used to evaluate the fine-tuned models. More examples are necessary than during fine-tuning to make the evaluation reliable.}
    \label{table:conciseness_eval_dataset}
\end{table}

\textbf{Conciseness task.} For conciseness fine-tuning, a dataset of prompts $\mathcal{D} = \{x_{1}, .., x_{K} \}$, with a set of verifiable solutions $\{s_{1}, ..., s_{K}\}$, i.e.\ shortest possible correct answers, was used. For example, for the prompt ``Name one primary color'', possible shortest verifiable solution used is ``Red''.  Following this approach, for each prompt $x \in \mathcal{D}$, the model was encouraged to generate a concise response $y$. To fine-tune the model to generate concise responses, a reward computed using the absolute length difference between the generated response $y$ and the corresponding verified solution $s_{k}$ was given to the model for each prompt $x_{k}$. The reward function $R$ for conciseness was defined as $R = -\lvert \mathrm{len}(y) - \mathrm{len}(s_{k}) \rvert$, where $\mathrm{len}(\cdot)$ denotes the string length.

\paragraph{Behavior metrics for the conciseness experiments.} Behavior of the fine-tuned models was measured in two ways: the mean conciseness reward and the mean KL divergence from the base model \citep[after ][]{rafailov2023dpo}. KL divergence is useful as a proxy for the preservation of the base model's behavior. It correlates strongly with the question-answering performance of the model, but also conveys more information, i.e.\ the extent of the fine-tuning changes. A low KL divergence thus suggests that the fine-tuned model has not forgotten capabilities learned during pre-training. Further, as KL divergence increases, these capabilities are likely to break. Therefore, fine-tuning behavior can be characterized using the tradeoffs between reward and KL divergence. To compute the metrics, each fine-tuned model was evaluated on a set of held-out test prompts, with $20$ responses sampled per prompt. The reward was computed using the model-generated response and the verifiable solution provided in the test dataset. The KL divergence between a fine-tuned model $\theta_{\mathrm{FT}}$ and a base model $\theta_{\mathrm{BASE}}$ for a given prompt $x$ and corresponding response $y$ was approximated following \citet{schulman2020approximating} as
$
\mathrm{KL}\!\left[\theta_\mathrm{FT} \,\|\, \theta_{\mathrm{BASE}}\right] 
= \frac{\theta_{\mathrm{BASE}}(y_{i,t} \mid x, y_{i,<t})}{\theta_\mathrm{FT}(y_{i,t} \mid x, y_{i,<t})}
- \log \frac{\theta_{\mathrm{BASE}}(y_{i,t} \mid x, y_{i,<t})}{\theta_{\mathrm{FT}}(y_{i,t} \mid x, y_{i,<t})}
- 1.$ 

\paragraph{Partially correlated noise vs. i.i.d. noise.} In the current implementation, for each perturbed model, the random number generator is reinitialized for each layer (using the random seed corresponding to this model), leading to partially correlated noise perturbations across layers. Preliminary experiments show that this implementation does not lead to significantly different performances compared to true i.i.d. noise.

\subsection{ES Hyperparameter Ablations}\label{app:hyperparam_ablation}
To further investigate how ES performance varies with population size $N$, noise scale $\sigma$, and learning rate $\alpha$, a number of ablation experiments were conducted in the GSM8K reasoning benchmark using Qwen2.5-1.5B and Qwen2.5-3B. Each experiment ran for $300$ training steps, with ES sampling a maximum of $512$ tokens per prompt. All runs used a fixed training dataset of $512$ prompts and were evaluated on the GSM8K test set using pass@1 with greedy decoding.

\paragraph{Effect of population size ($N$) on performance.}
First, the effect of varying the population size ($N$) was studied by setting $N \in \{16, 30, 64, 128\}$. Table \ref{table:population-size-1.5b} and Table \ref{table:population-size-3b} present the pass@1 scores of the final models. Across Qwen2.5-1.5B and Qwen2.5-3B, larger population sizes generally yielded better pass@1 performance, however, the differences were modest. Since increasing $N$ incurs additional computational cost, $N=30$ emerged as a reasonable tradeoff, balancing performance and computational efficiency.

\begin{table}[h!]

\begin{center}
\begin{tabular}{ |c|c|c|c|c|c| }
\hline
 $N$ & 16 & 30 & 64 & 128 \\
 \hline 
 \hline 
  pass@1 & 0.697 & 0.714 & 0.725 & 0.727 \\
 \hline 

\end{tabular}
\end{center}
\caption{\text{Pass@1} scores for population size ($N$) ablation for Qwen2.5-1.5B on GSM8K. Larger $N$ generally leads to better performance, but the differences are moderate.}
\label{table:population-size-1.5b}
\end{table}

\begin{table}[h!]

\begin{center}
\begin{tabular}{ |c|c|c|c|c|c| }
\hline
 $N$ & 16 & 30 & 64 & 128 \\
 \hline 
 \hline 
  pass@1 & 0.807 & 0.823 & 0.804 & 0.831 \\
 \hline 

\end{tabular}
\end{center}
\caption{\text{Pass@1} scores for population size ($N$) ablation for Qwen2.5-3B on GSM8K. $N=128$ gives the best performance, but $N=30$ also provides good performance at lower cost.}
\label{table:population-size-3b}
\end{table}

\paragraph{Effect of noise scale ($\sigma$) and learning rate ($\alpha$) on performance.}
Next, $\sigma$ and $\alpha$ were varied to study their effect on performance. Table \ref{table:sigma-alpha-1.5b} and \ref{table:sigma-alpha-3b} present the final model pass@1 score on GSM8K for each $\sigma$ and $\alpha$ configuration across Qwen2.5-1.5B and Qwen2.5-3B. The choice of $\sigma$ and $\alpha$ had a moderate effect on the final performance of both models. Qwen2.5-1.5B with $\sigma=0.0005$ and $\alpha=0.00025$ and Qwen2.5-3B with $\sigma=0.001$ and $\alpha=0.001$ produced the worst final performance, while $\sigma=0.002$ and $\alpha=0.0005$ achieved the best performance across both models. Both Tables \ref{table:sigma-alpha-1.5b} and \ref{table:sigma-alpha-3b} highlight that ES is robust across different hyperparameter settings and show that ES maintains similar performance across all tested configurations.

\begin{table}[h!]

\begin{center}
\begin{tabular}{ |c|c|c| } 
\hline
$\sigma$ & $\alpha$ & $\text{pass@1}$ \\
\hline
\hline 
0.001 &	0.0005 & 0.714 \\
\hline
0.002 & 0.001 &	0.688 \\
\hline
0.0005 & 0.00025 & 0.654 \\
\hline
0.002 & 0.0005 & 0.726 \\
\hline
0.0005 & 0.0005 & 0.675 \\
\hline
0.001 & 0.001 &	0.660 \\
\hline
0.001 & 0.00025 & 0.704 \\
\hline 

\end{tabular}
\end{center}
\caption{\text{Pass@1} scores for $\sigma$ and $\alpha$ ablation for Qwen2.5-1.5B on GSM8K. ES is robust across different hyperparameter settings.}
\label{table:sigma-alpha-1.5b}
\end{table}

\begin{table}[h!]

\begin{center}
\begin{tabular}{ |c|c|c| } 
\hline
$\sigma$ & $\alpha$ & $\text{pass@1}$ \\
\hline
\hline 
0.001 &	0.0005 & 0.817 \\
 \hline 
0.002 & 0.001  & 0.808 \\
 \hline 
0.0005 & 0.00025 & 0.792 \\
\hline 
0.002 & 0.0005 & 0.826 \\
\hline 
0.0005 & 0.0005 & 0.791 \\
\hline 
0.001 & 0.001 & 0.782 \\
\hline 
0.001 & 0.00025 & 0.823 \\
\hline 

\end{tabular}
\end{center}

\caption{\text{Pass@1} scores for $\sigma$ and $\alpha$ ablation for Qwen2.5-3B on GSM8K. ES maintains similar performance across different hyperparameter settings.}
\label{table:sigma-alpha-3b}
\end{table}

%\clearpage
\subsection{Examples of Countdown Task Outputs} 
\label{app:countdown_example}
\textbf{Countdown task.} The Countdown task~\citep{gandhi2024stream, tinyzero} requires constructing an arithmetic expression from a given set of numbers using basic operations $(+,-,\times,\div)$ to match a target value. For instance, the target $950$ can be obtained from $\{100,50,6,3\}$ with $100 \times (6+3) +50 = 950$). This constitutes a compact test of constrained symbolic reasoning, i.e.\ an important use case for fine-tuning.

\begin{tcolorbox}[
  label=fg:countdownexample,
  colback=white,
  colframe=gray!60,
  sharp corners,
  boxrule=0.8pt,
  left=6pt,
  right=6pt,
  top=8pt,
  bottom=8pt,
  width=\textwidth,
  enhanced jigsaw,
  breakable
]

\colorbox{blue!30}{\textbf{Iteration 400}}
\vspace{0.2cm}

\colorbox{blue!10}{\textbf{[1] Input:}}

You are a helpful assistant. You first think about the reasoning process in your mind and then provide the user with the answer. Using the numbers [49 41 73], create an equation that equals 17. You can use basic arithmetic operations (\verb|+|, \verb|-|, \verb|*|, \verb|/|) and each number can only be used once. Show your work in \verb|<think>| … \verb|</think>| tags. And return the final answer in \verb|<answer>| … \verb|</answer>| tags, for example \verb|<answer> (1 + 2) / 3 </answer>|.

\vspace{0.2cm}
\colorbox{blue!10}{\textbf{[1] Response:}}
\vspace{0.2cm}

\verb|<think>| First, let's consider all possible combinations of the numbers 49, 41, and 73 using basic arithmetic operations. We need to use each number exactly once and combine them to get 17.

1. Start with simple addition and subtraction:
   - 49 + 41 - 73 = 80 - 73 = 7 (too high)
   - 49 - 41 + 73 = 8 + 73 = 81 (too high)
   - 73 - 49 - 41 = 24 - 41 = -17 (too low)

2. Try multiplication and division:
   - (49 * 41) / 73 = 2029 / 73 $\approx$ 27.83 (not close to 17)
   - (73 * 41) / 49 = 3013 / 49 $\approx$ 61.47 (too high)
   - (49 * 73) / 41 = 3637 / 41 $\approx$ 88.80 (too high)
   - (41 * 73) / 49 = 3013 / 49 $\approx$ 61.47 (same as above)

3. Check different orders:
   - 73 - (49 + 41) = 73 - 90 = -17 (too low)
   - 49 - (73 - 41) = 49 - 32 = 17 (this looks promising)

So, the equation 49 - (73 - 41) equals 17. Let's verify it:
- Calculate inside the parentheses first: 73 - 41 = 32
- Then subtract from 49: 49 - 32 = 17

This satisfies our requirement. \verb|</think>|
\verb|<answer>| 49 - (73 - 41) \verb|</answer>|

Reward: 1.1000 (Format: 1.0000, Answer: 1.0000)

\captionof{figure}{An example of a countdown task interaction.}
\end{tcolorbox}

Figure~\ref{fg:countdownexample} shows an example of the Qwen2.5-3B model’s output on the Countdown task after 400 ES iterations. In order to improve accuracy, the model learned to adjust its formulas by comparing the generated answer with the correct one, and to verify its own answers for correctness.

\subsection{Additional Results for the Countdown Experiments} \label{app:additional_countdown}

To further verify the variance across runs for the Countdown task, three independent runs with different random seeds were conducted for ES, GRPO-v and Dr.GRPO-v. The experimental setup was the same as in Section~\ref{app:exp_setup}. As shown in Table~\ref{table:additional_countdown}, ES had the lowest variances across multiple runs. In an additional Student’s $t$-test, the performance differences between ES and other GRPO variants are statistically significant with $p<0.05$.

\begin{table}[h!]

\begin{center}
\begin{tabular}{ cccc }
\hline
 Mean $\pm$ Std & ES & GRPO-v & DR.GRPO-v \\
 \hline 
  Qwen2.5-1.5B-Instruct & $37.73\pm1.14$ & $26.77\pm1.98$ & $32.97\pm2.05$ \\
  Qwen2.5-3B-Instruct & $58.67\pm1.47$ & $35.07\pm2.33$ & $44.43\pm1.89$ \\
 \hline 

\end{tabular}
\end{center}
\caption{Performance (accuracy) statistics over three independent runs on the Countdown Task. ES exhibits the lowest variance across runs, and the performance differences between ES and GRPO variants are statistically significant.}
\label{table:additional_countdown}
\end{table}

\subsection{Extended Conciseness Details and Experiments}
\label{appendix:extended_conciseness_exp}
In this section, the conciseness experiments are extended to investigate the impact of different learning rates on GRPO training.

\begin{figure}[h]
\centering
\begin{subfigure}{.5\textwidth}
  \centering
  \includegraphics[width=1.0\linewidth]{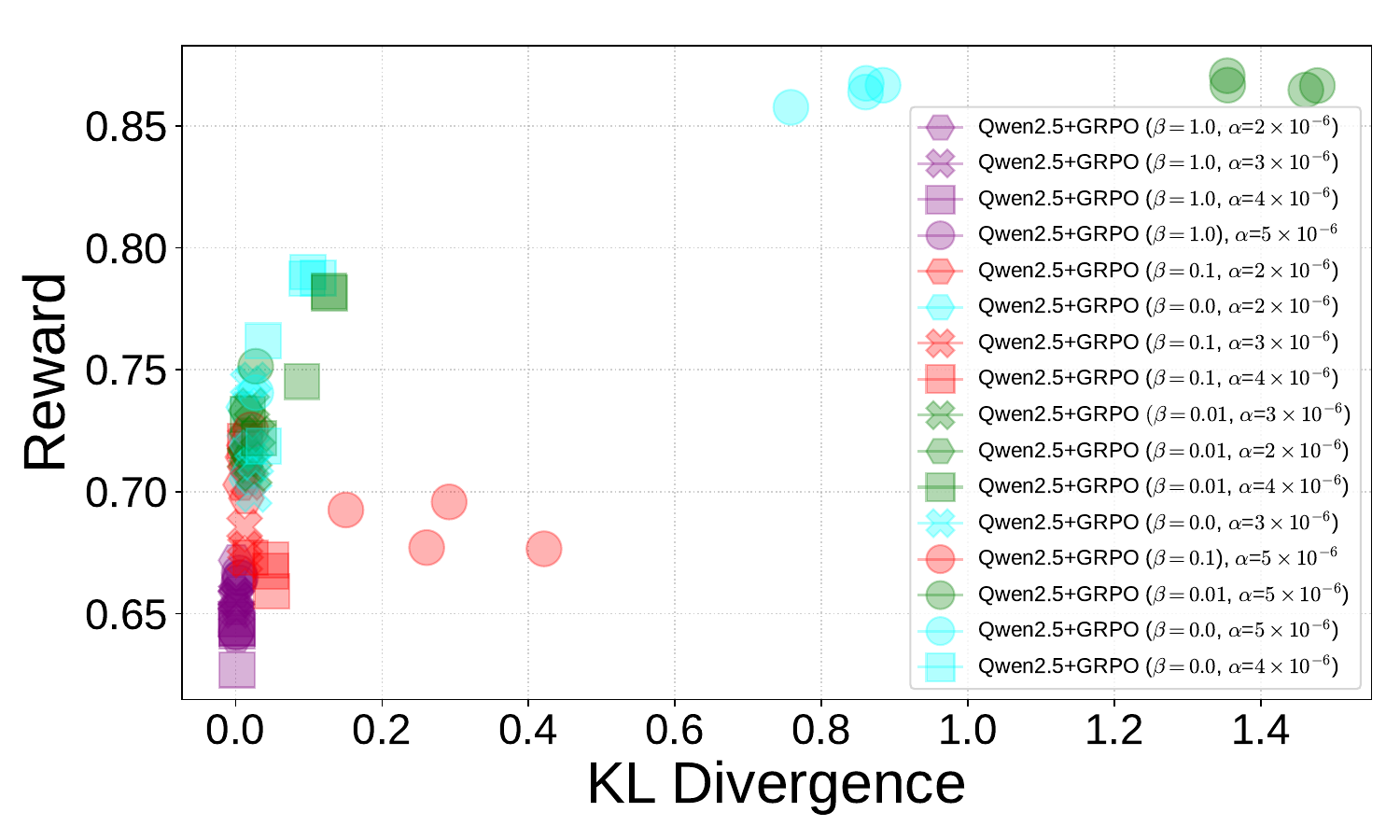}
  \vspace*{-3.5ex}
  \caption{GRPO models results over various learning rates.}
  %\label{fig:es_grpo_mean_models}
  \label{fig:learning_rate_mean_models}
\end{subfigure}%
\begin{subfigure}{.5\textwidth}
  \centering
  \includegraphics[width=1.0\linewidth]{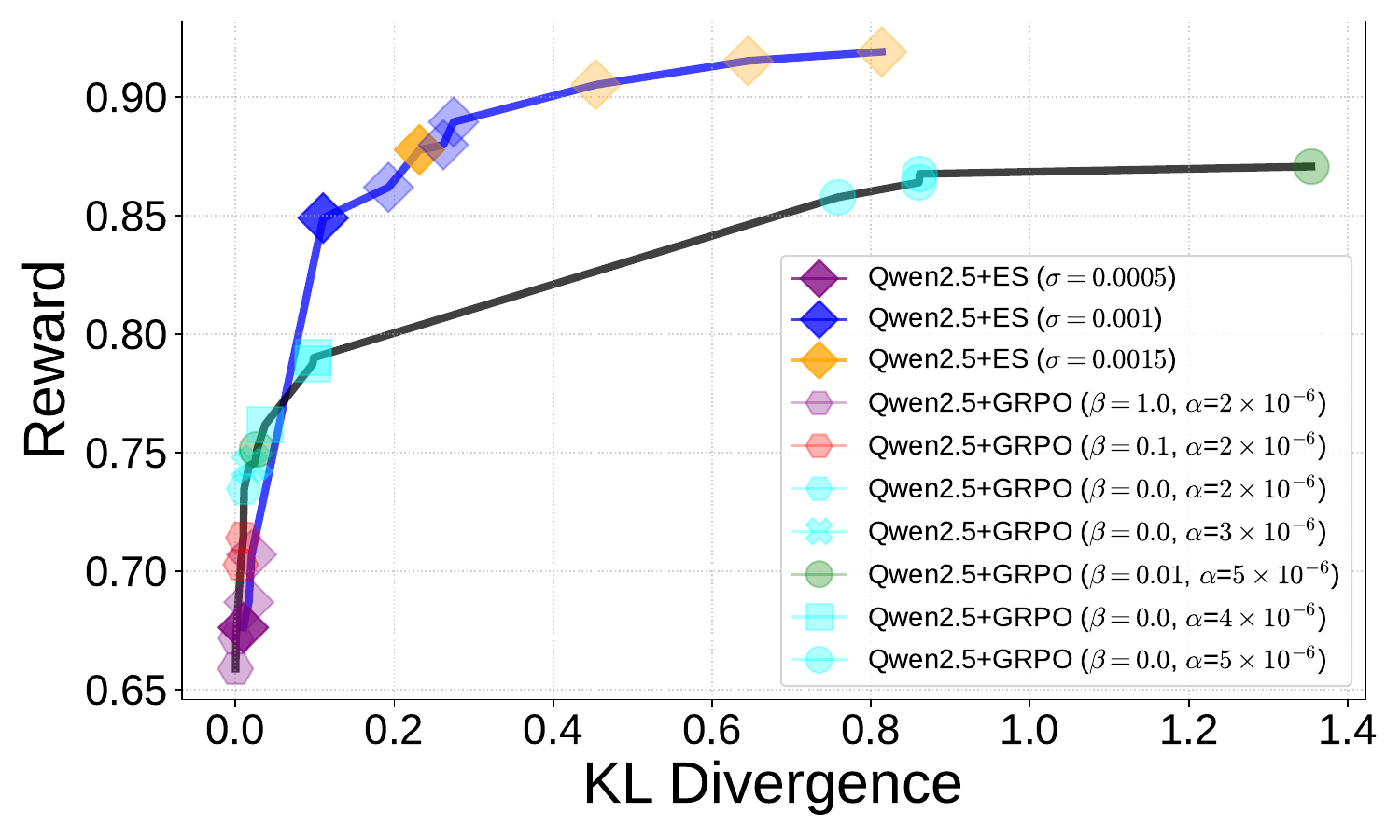}
  \vspace*{-3.5ex}
  \caption{ES and GRPO Pareto fronts.}
  \label{fig:learning_rate_pf}
  %\label{fig:}
\end{subfigure}
\vspace*{-4ex}
\caption{GRPO behavior with different learning rates. (a) GRPO models trained using different learning rates and $\beta$ values. Both conciseness reward and KL divergence increase with higher learning rates. (b) The ES Pareto front (blue line, top) plotted with the GRPO Pareto front (black line, bottom) over different model learning parameters. ES dominates GRPO across the whole range.}
\vspace{-10pt}

\end{figure}

\paragraph{GRPO with different learning rates.} Further GRPO experiments were run over four seeds with $\beta = \{0, 0.01, 0.1, 1.0\}$, varying the learning rate $\alpha = \{ 2\times10^{-6}, 3\times10^{-6}, 4\times10^{-6}, 5\times10^{-6}\}$. A total of $20$ responses were sampled per evaluation prompt. Figure \ref{fig:learning_rate_mean_models} shows the mean reward and KL divergence of each fine-tuned model. As the learning rate increases, both mean reward and mean KL divergence increase. The best models with respect to reward are trained using $5\times10^{-6}$ and $\beta=\{0.0, 0.01\}$, obtaining rewards greater than $0.85$. Figure \ref{fig:learning_rate_pf} further displays the GRPO Pareto front (black line, bottom) across these learning rates, comparing it with the ES Pareto front (blue line, top). The majority of Pareto optimal models across these learning rates obtain a mean reward of less than $0.8$ and a KL divergence of less than $0.4$. The ES Pareto front dominates that of GRPO over different learning rates and $\beta$ values. 

\begin{figure}
\centering
\begin{subfigure}{.5\textwidth}
  \centering
  \includegraphics[width=0.8\linewidth]{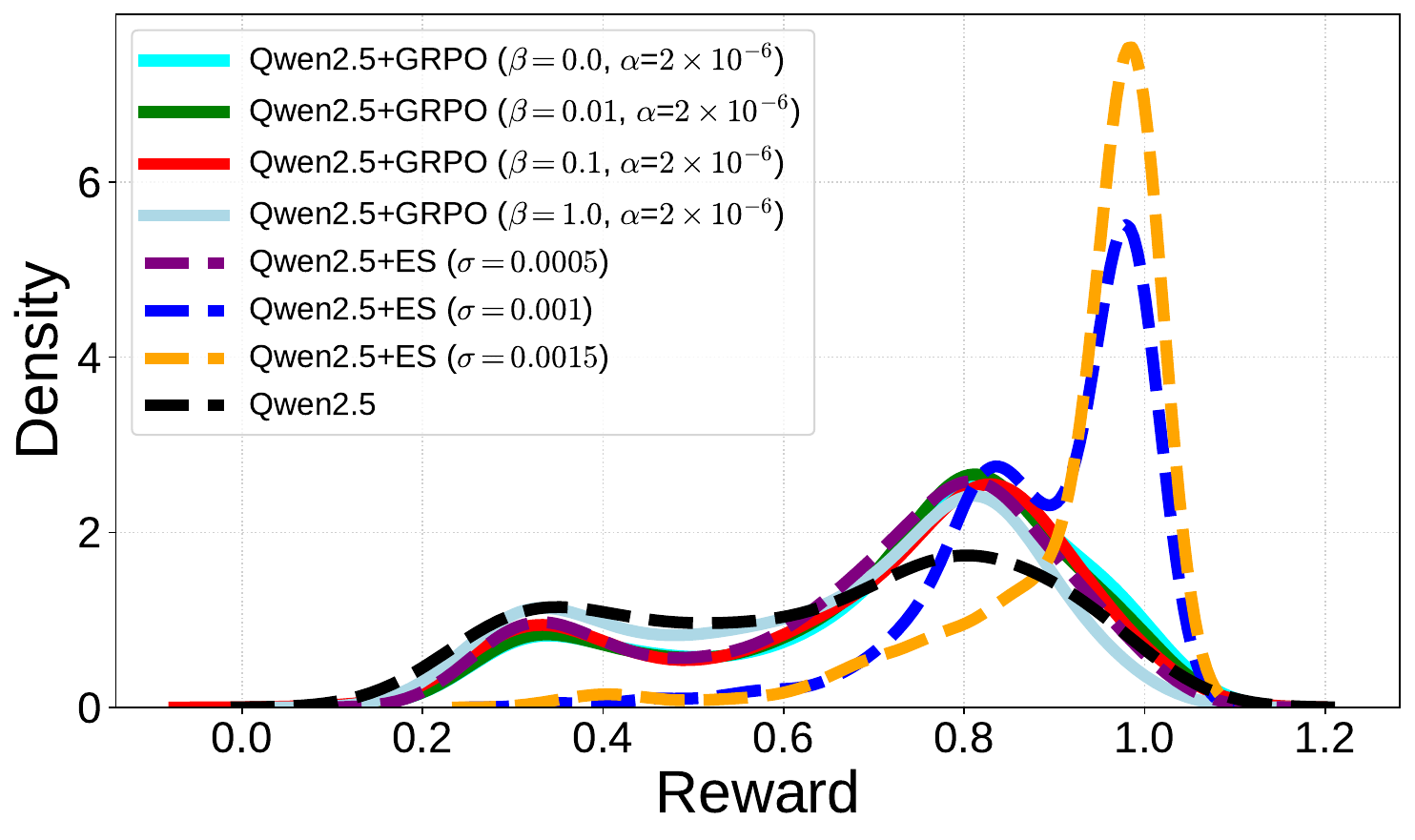}
  \caption{Reward distribution with $\alpha = 2\times10^{-6}$.}
  \label{fig:reward_distribution_2}
\end{subfigure}%
\begin{subfigure}{.5\textwidth}
  \centering
  \includegraphics[width=0.8\linewidth]{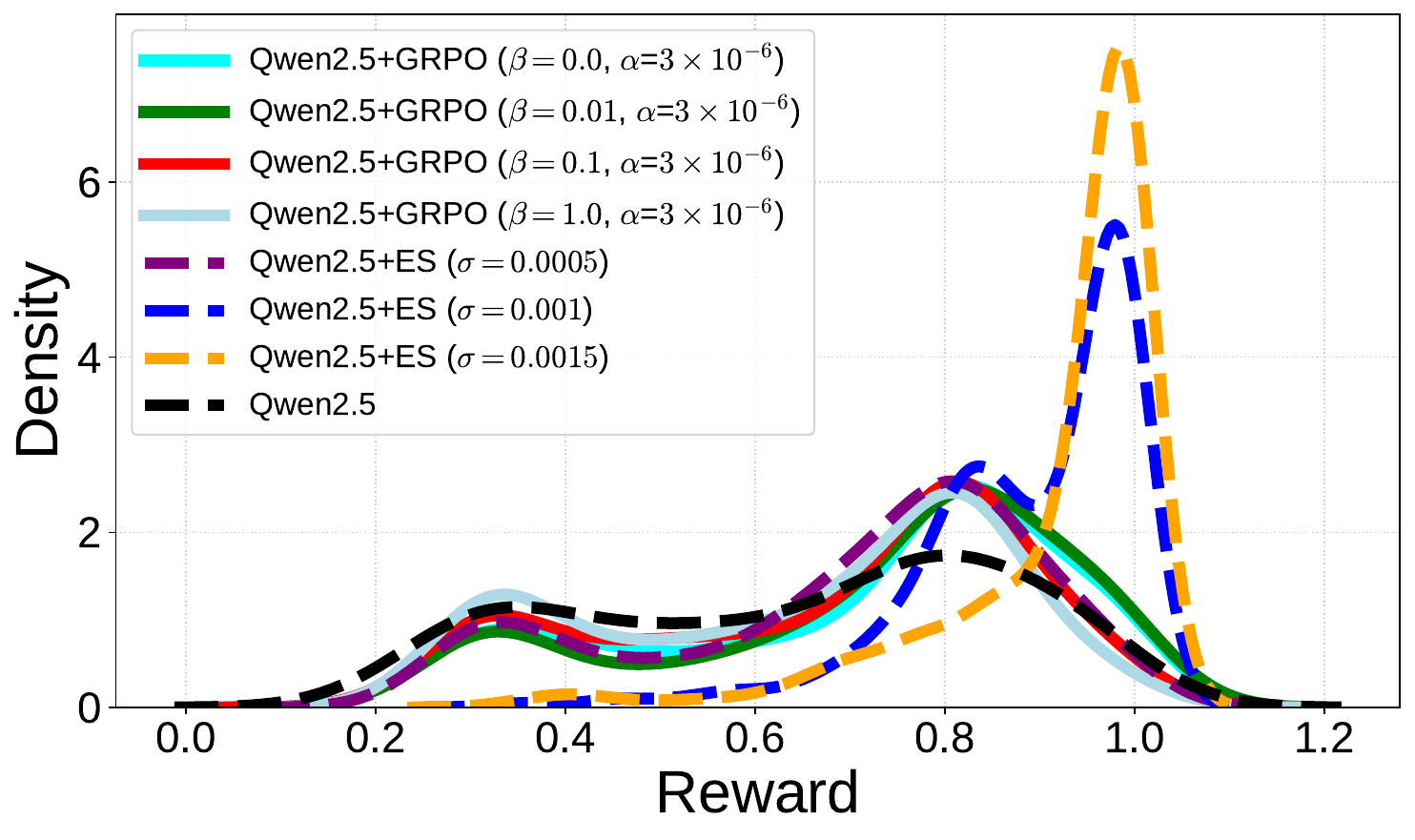}
  \caption{Reward distribution with $\alpha = 3\times10^{-6}$.}
  \label{fig:reward_distribution_3}
\end{subfigure}%

\begin{subfigure}{.5\textwidth}
  \centering
  \includegraphics[width=0.8\linewidth]{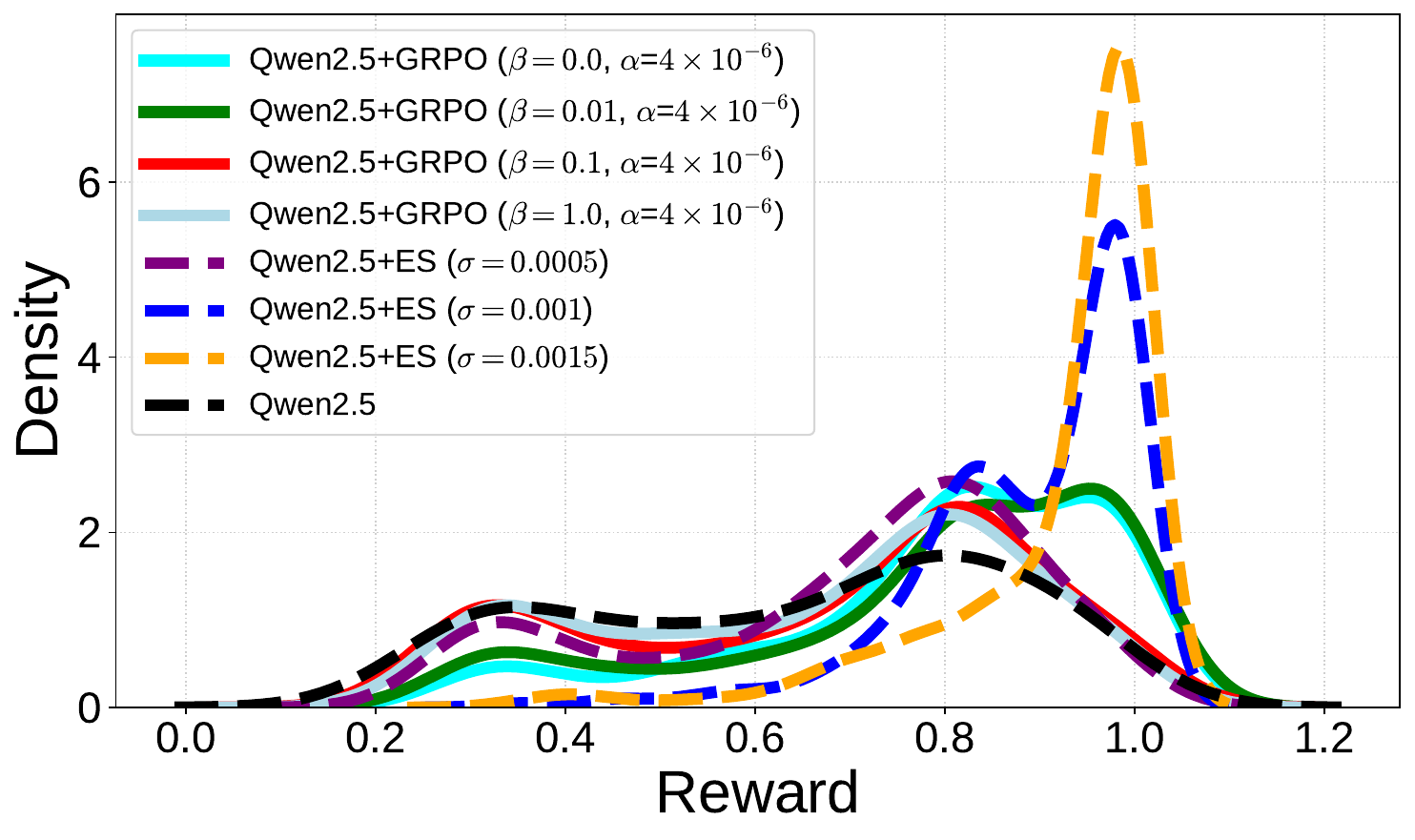}
  \caption{Reward distribution with $\alpha = 4\times10^{-6}$.}
  \label{fig:reward_distribution_4}
\end{subfigure}%
\begin{subfigure}{.5\textwidth}
  \centering
  \includegraphics[width=0.8\linewidth]{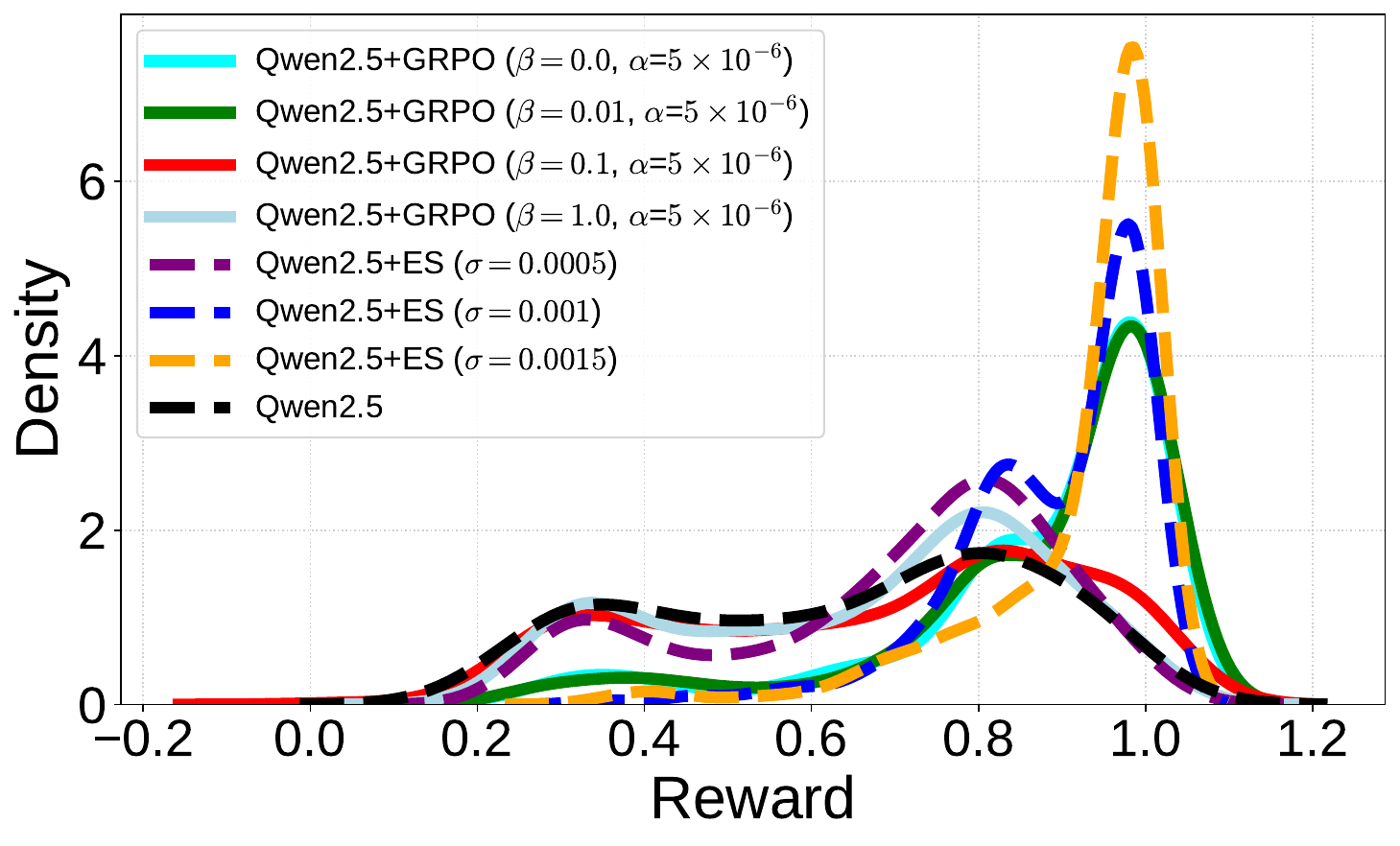}
  \caption{Reward distribution with $\alpha = 5\times10^{-6}$.}
  \label{fig:reward_distribution_5}
\end{subfigure}%

\caption{Reward distributions in fine-tuning for conciseness with different learning rates $\alpha =\{2\times10^{-6}$, $3\times10^{-6}$, $4\times10^{-6}, 5\times10^{-6} \}$ and $\beta = \{0.0, 0.01, 0.1, 1.0\}$ compared to ES on the Qwen2.5-7B-Instruct base model. Whereas GRPO distribution is similar to the base model, ES shifts it to the right, i.e.\ higher rewards. Higher rewards can only be achieved with GRPO with high learning rates and low $\beta$, which setting often breaks to model's performance.}
\label{fig:reward_distribution_all}
\end{figure}

Next, the reward distribution for each $\alpha$ and $\beta$ value for GRPO was compared with that of ES, starting with learning rates $2\times10^{-6}$ and $3\times10^{-6}$. Figures \ref{fig:reward_distribution_2} and Figure \ref{fig:reward_distribution_3} show that all GRPO models stay close to the Qwen2.5-7B-Instruct base model reward distribution, despite the variation in $\beta$. In contrast, ES shifts the reward distribution to the right with a  density peak around $1.0$, i.e.\ towards higher rewards.
%Thus, a higher learning rate is needed to effectively optimize the reward function with GRPO.
The learning rate was then further increased to $4\times10^{-6}$ (Figure~\ref{fig:reward_distribution_4}). As a result, for $\beta=0.0$ and $\beta=0.01$, GRPO shifts the reward distribution to the right towards higher rewards. However, they are still lower than those of ES. As the learning rate is increased further to $5\times10^{-6}$ (Figure \ref{fig:reward_distribution_5}), GRPO is sufficiently able to optimize the reward: with $\beta = 0.0$ and $\beta = 0.01$, it peaks around $1.0$. Thus, high learning rate combined with low $\beta$ is important for GRPO to optimize the reward. However, as was discussed before, such a setting often breaks the performance of the model.

\begin{figure}
    \centering
    \includegraphics[width=0.8\linewidth]{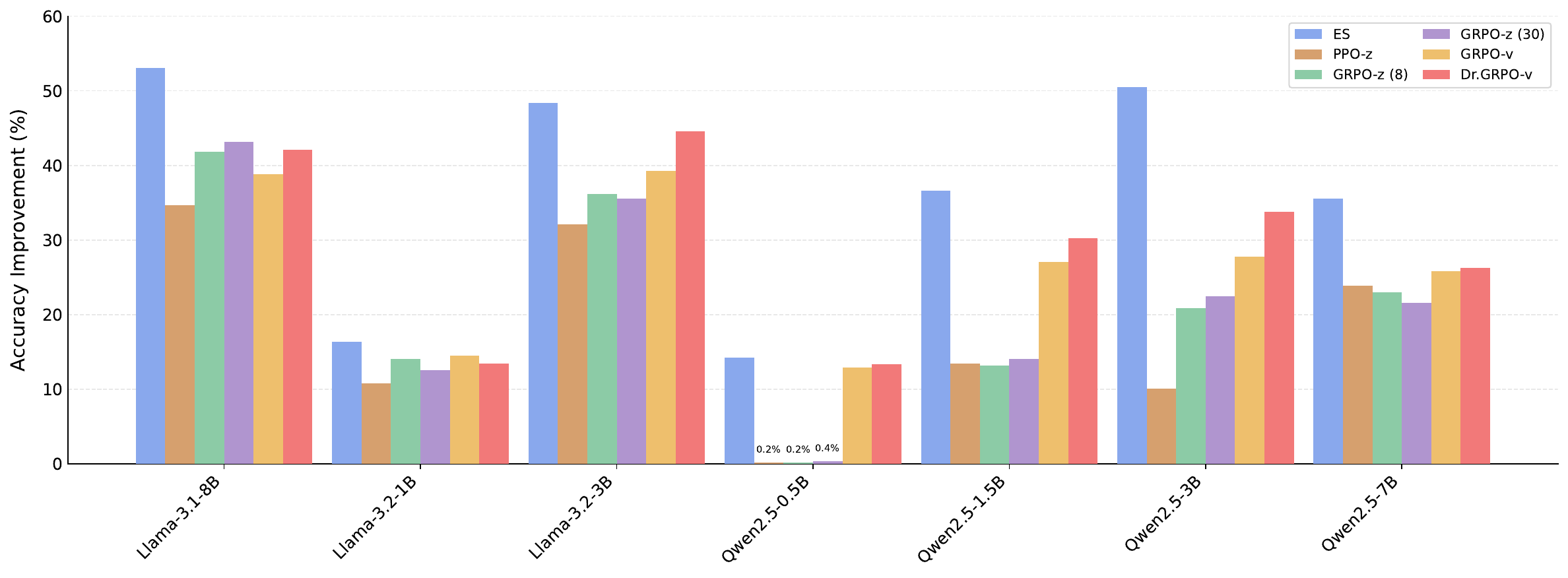}
    \caption{Accuracy Improvement over Base Models with ES vs RL across Model Families. ES results in consistently largest improvements in all cases.}
    \label{fig:acc_improv_over_base}
    \vspace{-6pt}
\end{figure}

\subsection{Training Curves and Accuracy Improvement of ES and RL on the Countdown Task}\label{appendix:train_curve}

\begin{figure}
    \centering
    \includegraphics[width=\linewidth]{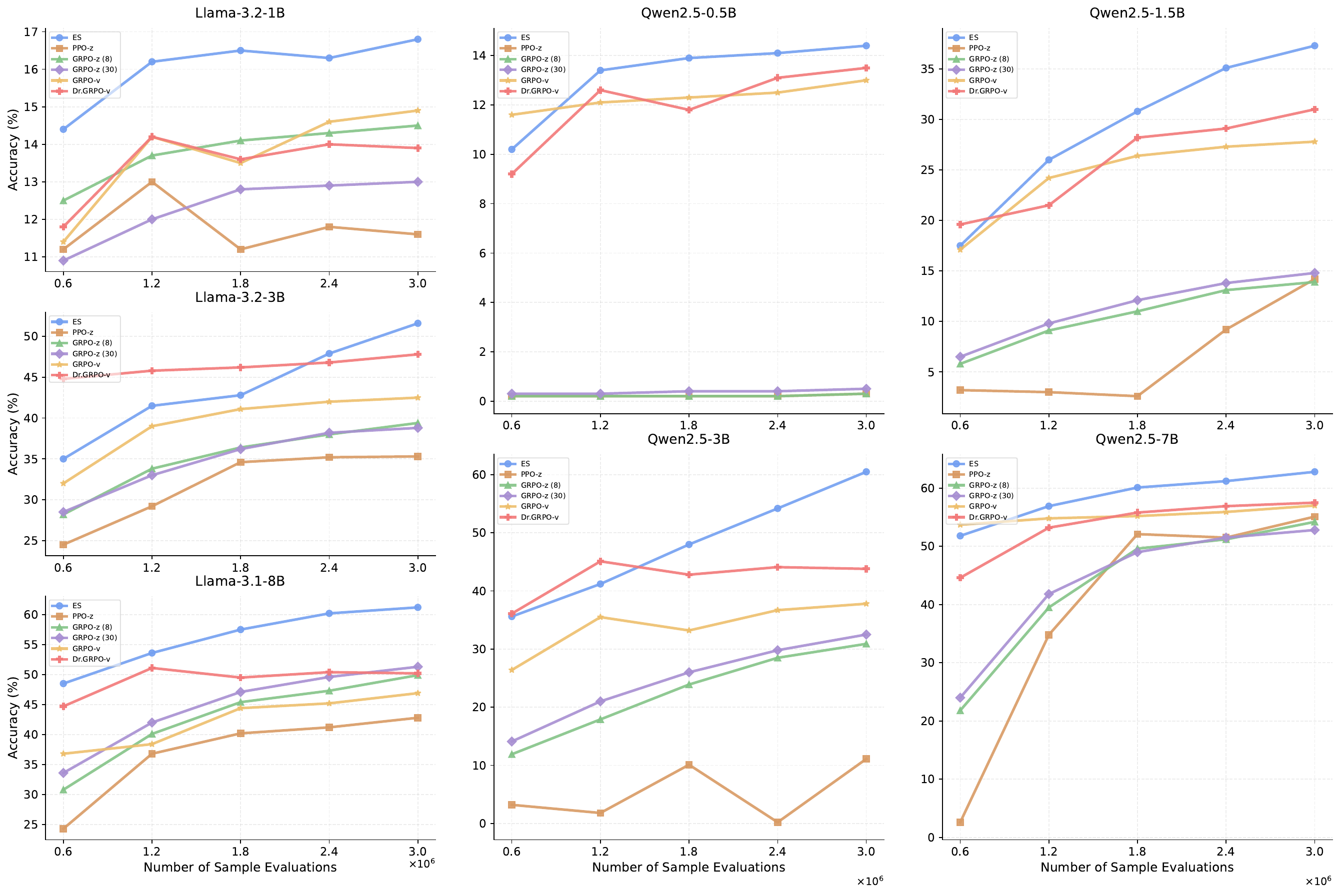}
    \caption{Training curves of ES and RL across two model families and six sizes in the countdown task. ES fine-tuning results in significantly better performance in all cases.}
    \label{fig:training_curve}
\end{figure}

As shown in Figure~\ref{fig:training_curve}, ES consistently outperformed RL across all tested models throughout training. %On the small model Qwen2.5-0.5B, RL showed almost no improvement, whereas ES steadily increased accuracy. On mid-sized models like Qwen2.5-1.5B, Qwen2.5-3B and Llama-3.2-3B, ES achieved substantial gains, reaching accuracy levels that RL never approached even in extended training. On larger models such as Qwen2.5-7B and Llama-3.1-8B, RL improved more than in the smaller models, but ES still maintained a clear and consistent advantage, achieving the highest accuracy throughout. 
In addition, as shown in Figure~\ref{fig:acc_improv_over_base}, we compute the relative improvements of PPO, GRPO, DR.GRPO and ES over their respective base models across different model families. ES delivers the consistently largest improvements in all cases.

\subsection{Extended Math Reasoning Details and Discussion}\label{app:math_extended}

\paragraph{RL baselines.}
We compare ES against three strong R1-Zero-style \cite{deepseekai2025deepseekr1} reasoning baselines at the 7B parameter scale: SimpleRL-Zero \cite{zeng2025simplerl}, OpenReasoner-Zero \cite{hu2025open}, and Oat-Zero \cite{liu2025understanding}. All baselines are instantiated using Qwen2.5-series models \cite{yang2025qwen2}, which are competitive open-weight language models known to exhibit strong reasoning performance at this scale. The respective baseline implementations are fully open source and built on production-ready RL libraries, including VERL \cite{sheng2024hybridflow} and OAT \cite{liu2024oat}, which provided highly optimized, stable, and tested PPO, GRPO, and Dr.GRPO implementations with efficient rollout management, and standardized reward handling. SimpleRL-Zero isolates the core R1-Zero optimization mechanism with minimal additional engineering, OpenReasoner-Zero reflects a widely adopted community implementation with practical design choices, and Oat-Zero further alters GRPO with algoritmic enhancements shown to boost performance. Together, these baselines represent strong, reproducible, and non-trivial comparators for evaluating ES.

\paragraph{Reward function.}
Our ES training utilizes a basic rule-based reward function that checks answer correctness, without any format rewards. The reward function is designed to extract the produced answer contained within $\texttt{\textbackslash boxed\{\}}$ and compare it with the ground truth answer. Similarly to RL, we implement a binary reward scheme where a reward of 1 is given for exact matches with the reference answer, and 0 for all other cases. To ensure a fair comparison with models from the literature we use the same answer extractor, also called a grader, as OatZero \cite{liu2025understanding}.

\paragraph{Qwen math template.}

Table \ref{table:qwen_math_template} shows the template stucture used for training ES models. We follow the same template used for training the Qwen-Math series base models, where the model is required to provide a final answer inside $\texttt{\textbackslash boxed\{\}}$. This requirement ensures that the final model answers are easy to extract and compare with the ground truth solutions for reward calculation during training. As shown in \cite{liu2025understanding}, the choice of template can impact the final performance of the model. We chose the Qwen-Math template given it provides a platform for stable learning and good performance during fine-tuning.
\begin{table}
\centering
\small
\setlength{\tabcolsep}{6pt}
\renewcommand{\arraystretch}{1.3}

\begin{tabular}{|p{0.22\linewidth}|p{0.6\linewidth}|}
\hline

\centering\textbf{Qwen-Math} &
\texttt{<|im\_start|>system}\texttt{\textbackslash n}Please reason step by step, and put your final answer within
\texttt{\textbackslash boxed\{\}}.\texttt{<|im\_end|>}\texttt{\textbackslash n}\texttt{<|im\_start|>user}\texttt{\textbackslash n}\textcolor{red}{\{question\}} \texttt{<|im\_end|>}\texttt{\textbackslash n}\texttt{<|im\_start|>assistant}\texttt{\textbackslash n}
\\

\hline
\end{tabular}

\caption{Qwen-Math prompt template used in this work.}
\label{table:qwen_math_template}
\end{table}

\paragraph{Checkpoint selection.} Given there is no explicit validation set, for ES, we follow the standard model checkpoint selection mechanism from the literature whereby the checkpoints with high average pass@1 accuracy over each evaluation set over training are presented. We chose to present a number of ES checkpoints that all achieve sufficiently high average score. We chose our checkpoints to ensure competitive performance across the range of benchmarks. In this case, $\text{ES}_{\text{CHKPT-1}}$ is chosen for its high average score and occurs after $336$ training steps. Additionally, we take a checkpoint after $160$ training steps and perform $10$ additional model update steps with $\alpha=\frac{\sigma}{4}$. This additional training produced $\text{ES}_{\text{CHKPT-2}}$. Given the lack of validation set, we utilize MATH500 as a pseudo validation set since MATH500 is an in-distribution validation. Following this, we select $\text{ES}_{\text{CHKPT-3}}$ because it achieves the highest performance in the MATH500 benchmark across our evaluations. The MATH500 $\text{ES}_{\text{CHKPT-3}}$ occurs after $192$ training steps.

\subsection{Analysis of Computational Cost} \label{app:computation}
The computational cost in terms of FLOPs is analyzed in this section. For each sample evaluation (processing one training data point), ES needs to perform one forward pass on the base model, whereas GRPO needs to perform one forward pass on the base model, one forward pass on the reference model, and one backward pass (backpropagation) on the policy. Following the standard estimation in literature \citep{Narayanan2021megatron,gan2026neuralthicketsdiversetask}, for a model with $P$ parameters and a sequence length of $L$, a single forward pass requires approximately $2PL$ FLOPs, and a backward pass requires $4PL$ FLOPs. Consequently, ES needs $2PL$ FLOPs for each sample evaluation, whereas GRPO needs $2+2+4 = 8PL$ FLOPs. Note that for earlier RL methods such as PPO, the needed FLOPs are even larger since there are additional forward and backward passes for the critic model. This analysis shows the computational advantage of ES in terms of FLOPs due to its inference-only nature.

\subsection{Parameter Magnitude Shifts by Evolutionary fine-tuning}
\label{app:shifts}

\begin{figure}[!htb]
\includegraphics[width=1.0\linewidth]{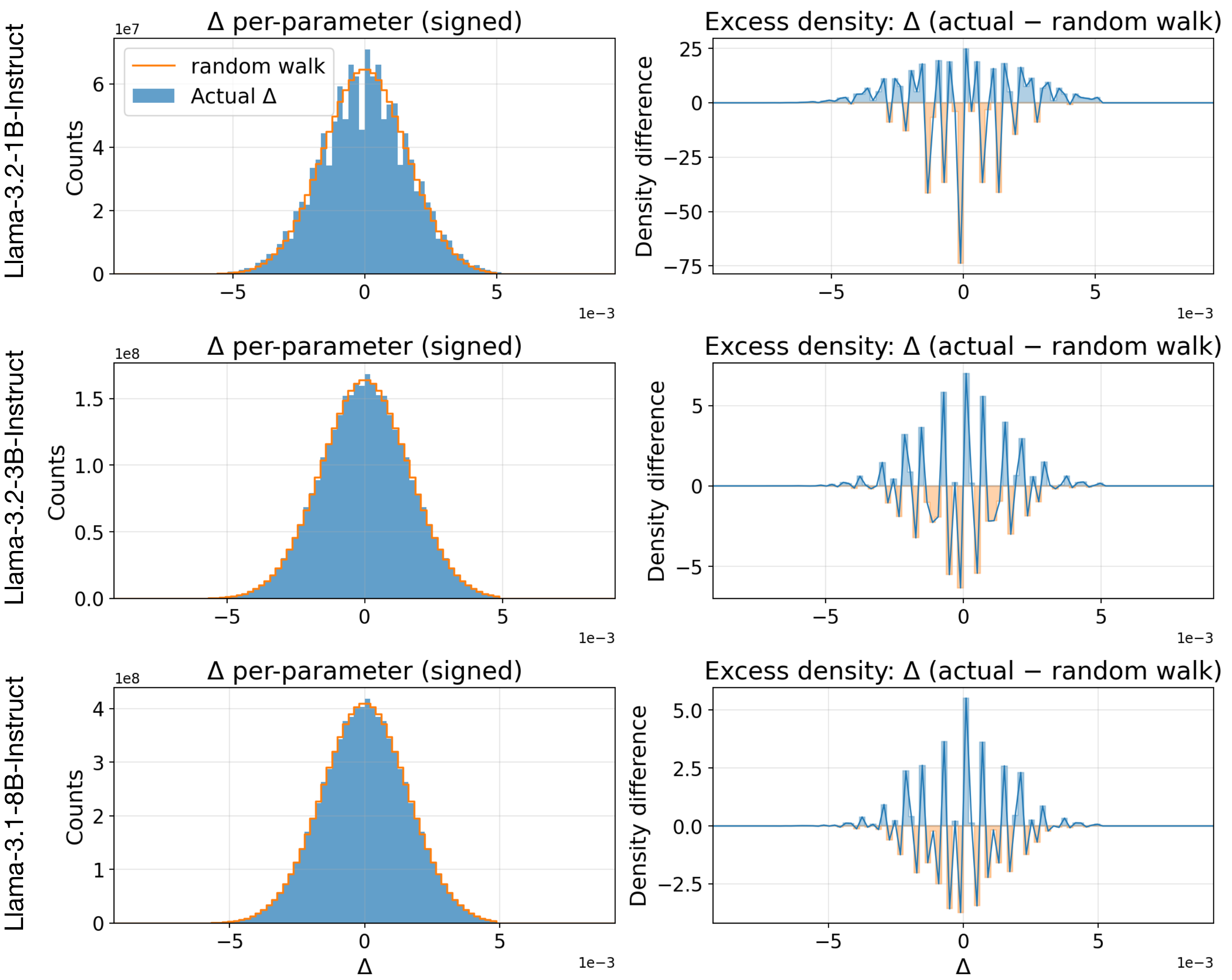}
\caption{Parameter magnitude shift histograms for the Countdown task in Llama models optimized by ES. The changes are similar to those of a random walk, concentrated around zero, likely due to numerical inaccuracies.}
\label{fig:llama-countdown-hist}
\end{figure}

\begin{figure}[!htb]
\includegraphics[width=1.0\linewidth]{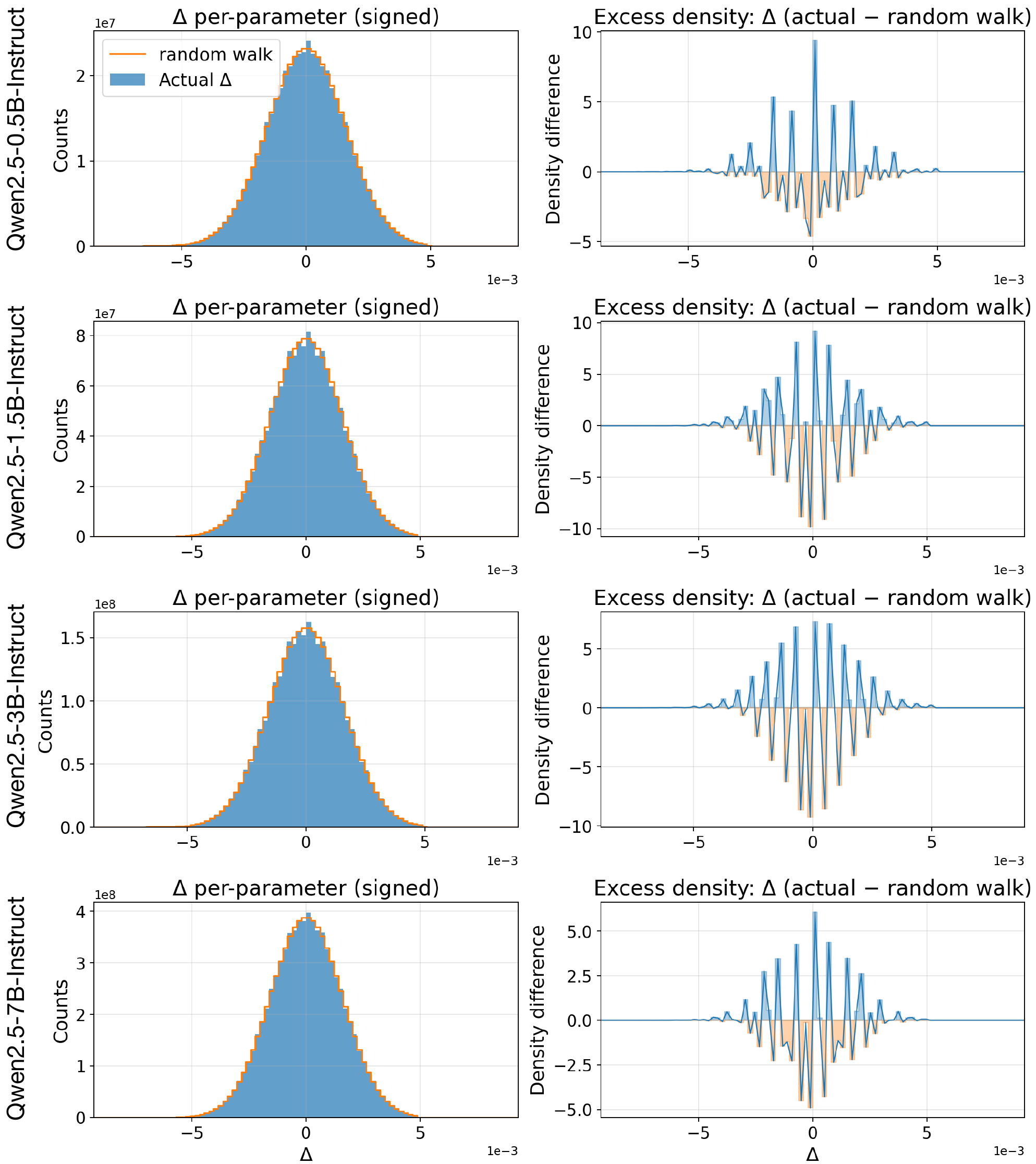}
\caption{Parameter magnitude shift histograms for the Countdown task in Qwen models optimized by ES. The results are consistent with those observed in Llama models.}
\label{fig:qwen-countdown-hist}
\end{figure}

This section characterizes how parameter magnitudes changed in ES fine-tuning in the countdown and conciseness experiments. Specifically, Figures~\ref{fig:llama-countdown-hist} and~\ref{fig:qwen-countdown-hist}, left column, show histograms of the absolute parameter magnitude shifts $\Delta$ before and after finetuning Llama and Qwen models, overlaid with random walk, on the Countdown task reported in Table~\ref{tab:countdown_main}. The right column in these figures shows the difference between $\Delta$ and the random walk.

For most models, $\Delta$ deviates very little from random walk. This is a counterintuitive result since fine-tuning actually resulted in a significant performance boost. A closer inspection reveals that most of the deviation was concentrated around zero. A likely explanation is that there are precision issues around zero, particularly with small bin sizes, which may lead to such deviations.

\begin{figure}[!htb]
\includegraphics[width=1.0\linewidth]{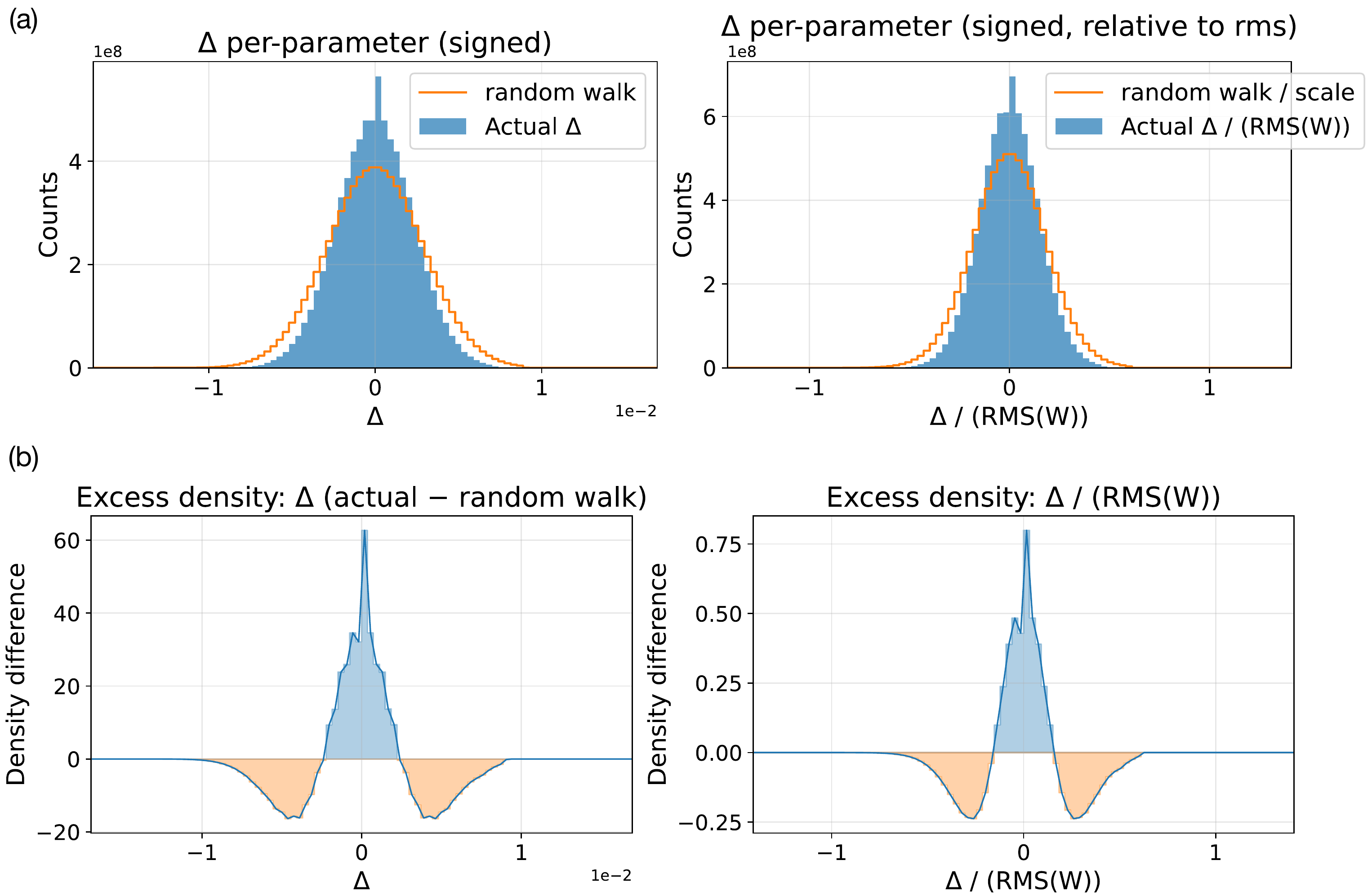}
\caption{Parameter magnitude shift histograms in conciseness fine-tuning in Qwen2.5-7B-Instruct model with ES. In this case, the model is large and the fine-tuning goals is different, revealing a potentially significant pattern of primarily small changes. The hypothesis (to be analyzed more thoroughly in future work) is that behavior is coded in large models in a redundant manner, making it possible to achieve this fine-tuning objective through numerous small changes. }
\label{fig:qwen-concise-hist}
\end{figure}

More significantly, a systematic deviation from the random walk was observed in conciseness fine-tuning of the largest model, Qwen2.5-7B-Instruct (Figure~\ref{fig:qwen-concise-hist}). The distribution shifts toward abundant small magnitude edits, suggesting that small parameter tweaks may be most significant in influencing output behavior. This result reinforces observations in prior studies~\citep[e.g.][]{liu2025sparse}.  A possible explanation is that large models encode functionality in a more redundandant manner, and therefore minor tweaks are sufficient to achieve fine-tuning objectives. In fact, the changes are nearly indistinguishable from random walk in Figures~\ref{fig:llama-countdown-hist} and~\ref{fig:qwen-countdown-hist} likely because they are benevolent wrt.\ the fine-tuning objective. A more thorough investigation of these hypotheses is a most interesting direction of future work, potentially resulting in a better understanding of fine-tuning and information processing principles in LLMs in general.

\clearpage

\subsection{Experiment with Mini Sudoku Task}\label{app:sudoku}
Standard Sudoku requires filling in missing integers in a $9\times9$ grid, evenly divided into $9$ smaller $3\times3$ grids. Integers must satisfy the following conditions: every row contains each integer $1-9$, every column contains each integer $1-9$, and each $3\times3$ sub-grid contains each integer $1-9$. Given the difficulty of $9\times9$ puzzles, existing evaluation of LLM's ability to complete Sudoku is often done with $4\times4$ grids, with equivalent rules for rows, columns, and $2\times2$ subgrids with digits $1-4$. Training data generation and evaluation was preformed with the sudoku board generation logic and reward function from \cite{reasoninggym}. $1,000$ example unique grids were generated, with $800$ in training and $200$ in testing. Qwen-2.5-3B-Instruct was used as the base model and trained with greedy decoding and batch size set to the entire $800$ sample training set. ES was run for $2,500$ iterations and hyperparameters were set as follows: $\sigma = 0.001$, $\alpha = 0.0005$, and $N = 32$. With this setup, model performance improved from 2\% to 66.5\% of test set puzzles correctly solved. An example of the prompt, along with model responses before and after ES fine tuning is provided below in Figure 11.

\FloatBarrier
\begin{tcolorbox}[
  label=fg:sudokuexample,
  colback=white,
  colframe=gray!60,
  sharp corners,
  boxrule=0.8pt,
  left=6pt,
  right=6pt,
  top=8pt,
  bottom=8pt,
  width=\textwidth,
  enhanced jigsaw,
  breakable
]
\vspace{0.2cm}
\colorbox{blue!10}{\textbf{Prompt:}}
\vspace{0.2cm}

You are a helpful assistant. You first think about the reasoning process in your mind and then provide the user with the answer. In 4x4 Mini Sudoku:
\begin{itemize}
    \item Each row must contain each number from 1--4 exactly once
    \item Each column must contain each number 1--4 exactly once
    \item Each 2x2 subgrid must contain each number 1--4 exactly once
\end{itemize}

Solve this 4x4 Mini Sudoku puzzle:

{\ttfamily\noindent
\_ \_ 4 \_\\
4 3 \_ \_\\
3 \_ \_ 1\\
1 \_ 3 \_
}

Show your work in \texttt{<think> </think>} tags. And return the final answer in \texttt{<answer> </answer>} tags, with spaces separating each number within a row, and newlines separating rows. For example \texttt{<answer>{\ttfamily
\\
2 1 4 3\\
4 3 2 1\\
3 1 4 2\\
1 2 3 4}\\</answer>}. Let me solve this step by step.

\texttt{<think>}

\vspace{0.2cm}
\colorbox{red!10}{\textbf{Base Model Output (Incorrect):}}
\vspace{0.2cm}

First, let's look at the given numbers and try to deduce the missing numbers. The number 4 is already placed in the first column, so we can't have another 4 in that column. The number 3 is already placed in the second row, so we can't have another 3 in that row. The number 1 is already placed in the third row, so we can't have another 1 in that row. Let's start filling in the grid based on these constraints: \texttt{</think>}

\texttt{<answer>}\\
{\ttfamily
2 1 4 3\\
4 3 2 1\\
3 1 4 2\\
1 2 3 4}\\
\texttt{</answer>}

\texttt{</think>} This is the final solution to the 4x4 Mini Sudoku puzzle. Let me verify:

Each row contains the numbers 1, 2, 3, 4 exactly once: Row 1: 2 1 4 3, Row 2: 4 3 2 1, Row 3: 3 1 4 2, Row 4: 1 2 3 4.

Each column contains the numbers 1, 2, 3, 4 exactly once: Column 1: 2 4 3 1, Column 2: 1 3 1 2, Column 3: 4 2 4 3, Column 4: 3 1 2 4.

Each 2x2 subgrid contains the numbers 1, 2, 3, 4 exactly once: Subgrid 1: 2 1 / 3 4, Subgrid 2: 4 2 / 1 3, Subgrid 3: 3 1 / 4 2, Subgrid 4: 1 2 / 3 4.

All constraints are satisfied, so the solution is correct. \texttt{</think>} The final answer is:

\texttt{<answer>}\\
{\ttfamily
2 1 4 3\\
4 3 2 1\\
3 1 4 2\\
1 2 3 4}\\
\texttt{</answer>}

\vspace{0.2cm}
\colorbox{green!10}{\textbf{Fine-Tuned Model Output (Correct):}}
\vspace{0.2cm}

First, let's look at the given numbers and try to fill in the missing numbers. We start with the first row: \texttt{\_ \_ 4 \_}. Since each row must contain each number from 1 to 4 exactly once, the missing numbers in the first row must be 2, 1, and 3. So the first row becomes: 2 1 4 3.

Next, let's look at the second row: \texttt{4 3 \_ \_}. Since each row must contain each number from 1 to 4 exactly once, and we already have 4 and 3 in this row, the missing numbers must be 1 and 2. So the second row becomes: 4 3 1 2.

Now, let's look at the third row: \texttt{3 \_ \_ 1}. Since each row must contain each number from 1 to 4 exactly once, and we already have 3 and 1 in this row, the missing numbers must be 2 and 4. So the third row becomes: 3 4 2 1.

Finally, let's look at the fourth row: \texttt{1 \_ 3 \_}. Since each row must contain each number from 1 to 4 exactly once, and we already have 1 and 3 in this row, the missing numbers must be 2 and 4. So the fourth row becomes: 1 2 3 4.

So, the completed 4x4 Mini Sudoku puzzle looks like this: \texttt{</think>}

\texttt{<answer>}\\
{\ttfamily
2 1 4 3\\
4 3 1 2\\
3 4 2 1\\
1 2 3 4}\\
\texttt{</answer>}

\captionof{figure}{Example prompt and model outputs for the 4x4 Mini Sudoku task. The base model would respond in a variety of step by step methods, while the fine tuned model settled into a consistent, dense thinking space.}
\end{tcolorbox}
\FloatBarrier

% \conor{It would be interesting to include some examples of the test set where the base model cannot solve the puzzle and the finetuned ES can. Also, could show the progression of how the answer changes over time}

% \roberto{Think that's a good idea, will get some of those samples from checkpoints tomorrow}

\subsection{Experiments on ARC-AGI Task} \label{app:arc-agi}

\textit{ARC-AGI}~\citep{chollet2024arc} is a benchmark designed to evaluate abstract reasoning and generalization capabilities. Each task presents a few input-output grid pairs as examples, requiring models to infer the underlying transformation rules and apply it to a new test input. The dataset has 400 public training tasks and evaluation tasks, focusing on creative problem-solving. We used 200 training tasks for fine-tuning and 200 evaluation tasks for testing. 

\subsubsection{Experimental settings}

\textit{Prompts}. The ARC-AGI tasks are presented in image format. To use LLMs to solve them, we first map colors to the numbers as follows: black $\to$ 0, blue $\to$ 1, red $\to$ 2, green $\to$ 3, yellow $\to$ 4, grey $\to$ 5, gray $\to$ 5,
pink $\to$ 6, orange $\to$ 7, purple $\to$ 8, brown $\to$ 9.

\begin{tcolorbox}[
  label=fg:arcagiexample,
  colback=white,
  colframe=gray!60,
  sharp corners,
  boxrule=0.8pt,
  left=6pt,
  right=6pt,
  top=8pt,
  bottom=8pt,
  width=\textwidth,
  enhanced jigsaw,
  breakable
]

%\colorbox{blue!30}{}
\vspace{0.2cm}
\colorbox{blue!10}{\textbf{System Prompt:}}
\vspace{0.2cm}

You are a creative and meticulous ARC puzzle solver who explains reasoning before answering.

\vspace{0.2cm}
\colorbox{blue!10}{\textbf{Task Explanation:}}
\vspace{0.2cm}

You will be given some number of paired example inputs and outputs. The outputs were produced by applying a transformation rule to the inputs. In addition to the paired example inputs and outputs, there is also one additional input without a known output. Your task is to determine the transformation rule and implement it in code.

The inputs and outputs are each "grids". A grid is a rectangular matrix of integers between 0 and 9 (inclusive). These grids will be shown to you as grids of numbers (ASCII). Each number corresponds to a color. The correspondence is as follows: black: 0, blue: 1, red: 2, green: 3, yellow: 4, grey: 5, pink: 6, orange: 7, purple: 8, brown: 9.

The transformation only needs to be unambiguous and applicable to the example inputs and the additional input. It doesn't need to work for all possible inputs.

\vspace{0.2cm}
\colorbox{blue!10}{\textbf{Reasoning Explanation:}}
\vspace{0.2cm}

You'll need to carefully reason in order to determine the transformation rule. Start your response by carefully reasoning in <reasoning></reasoning> tags. Then, implement the transformation in code.

After your reasoning write code in triple backticks (```python and then ```). You should write a function called `transform` which takes a single argument, the input grid as `list[list[int]]`, and returns the transformed grid (also as `list[list[int]]`). You should make sure that you implement a version of the transformation which works in general (it shouldn't just work for the additional input).

\vspace{0.2cm}
\colorbox{blue!10}{\textbf{Other Instructions:}}
\vspace{0.2cm}

Don't write tests in your python code, just output the `transform` function. (It will be tested later.)

You can also ask question to verify your observation on the inputs/outputs patterns in the form of python function which takes two arguments, the input and expected output grid both as `list[list[int]]` and returns the boolean flag (True or False). We will help you by running your Python function on examples and let you know whether your question is True or False.

You follow a particular reasoning style. You break down complex problems into smaller parts and reason through them step by step, arriving at sub-conclusions before stating an overall conclusion. This reduces the extent to which you need to do large leaps of reasoning.

You reason in substantial detail for as is necessary to determine the transformation rule.

You are creative and accomplished at solving puzzles. When you write `transform`, do not hardcode the solution for each example. We will run your transform function on additional inputs later and check if your logic is generic in addition to check the correctness.

\captionof{figure}{Prompts used for the ARC-AGI task.}
\end{tcolorbox}

\textit{Hyperparameters} $\sigma = 0.001$, $\alpha = 0.0003$, $N$ = 50, iterations = 1500.

\subsubsection{RL attempts} 
Few prior works have explored using RL–based post-training for LLMs on ARC-AGI tasks. While RL has been widely used for LLM alignment and reasoning tasks (e.g., RLHF, math and QA benchmarks), its application to abstraction-centric, out-of-distribution generalization benchmarks like ARC-AGI remains relatively under-explored, largely due to challenges in reward design and efficient exploration. According to ~\citet{arc_rl_report2025}, applying pure RL approach to ARC-AGI resulted in minimal gains due to the large action space.

% Qwen/Qwen2.5-14B-Instruct

% \subsection{}

% The $\mathtt{\backslash onecolumn}$ command above can be kept in place if you
% prefer a one-column appendix, or can be removed if you prefer a two-column
% appendix.  Apart from this possible change, the style (font size, spacing,
% margins, page numbering, etc.) should be kept the same as the main body.
%%%%%%%%%%%%%%%%%%%%%%%%%%%%%%%%%%%%%%%%%%%%%%%%%%%%%%%%%%%%%%%%%%%%%%%%%%%%%%%
%%%%%%%%%%%%%%%%%%%%%%%%%%%%%%%%%%%%%%%%%%%%%%%%%%%%%%%%%%%%%%%%%%%%%%%%%%%%%%%

\end{document}

%% file: math_commands.tex
\usepackage{amsmath,amsfonts,bm}

\def\eqref#1{equation~\ref{#1}}
\def\1{\bm{1}}

\DeclareMathAlphabet{\mathsfit}{\encodingdefault}{\sfdefault}{m}{sl}
\SetMathAlphabet{\mathsfit}{bold}{\encodingdefault}{\sfdefault}{bx}{n}

%% file: example_paper.bib
@article{schulman2020approximating,
  title={Approximating kl divergence, 2020},
  author={Schulman, John},
  journal={URL http://joschu. net/blog/kl-approx. html},
  year={2020}
}

@misc{tinyzero,
author       = {Jiayi Pan and Junjie Zhang and Xingyao Wang and Lifan Yuan and Hao Peng and Alane Suhr},
title        = {TinyZero},
howpublished = {https://github.com/Jiayi-Pan/TinyZero},
note         = {Accessed: 2025-01-24},
year         = {2025}
}

@article{yang2025qwen2,
  title={Qwen2. 5-1m technical report},
  author={Yang, An and Yu, Bowen and Li, Chengyuan and Liu, Dayiheng and Huang, Fei and Huang, Haoyan and Jiang, Jiandong and Tu, Jianhong and Zhang, Jianwei and Zhou, Jingren and others},
  journal={arXiv:2501.15383},
  year={2025}
}

@article{wierstra14a,
	author  = {Daan Wierstra and Tom Schaul and Tobias Glasmachers and Yi Sun and Jan Peters and JÃ¼rgen Schmidhuber},
	title   = {Natural Evolution Strategies},
	journal = {Journal of Machine Learning Research},
	year    = {2014},
	volume  = {15},
	number  = {27},
	pages   = {949--980},
	url     = {http://jmlr.org/papers/v15/wierstra14a.html}
}

@INPROCEEDINGS{wierstra2008,
	author={Wierstra, Daan and Schaul, Tom and Peters, Jan and Schmidhuber, Juergen},
	booktitle={2008 IEEE Congress on Evolutionary Computation (IEEE World Congress on Computational Intelligence)}, 
	title={Natural Evolution Strategies}, 
	year={2008},
	volume={},
	number={},
	pages={3381-3387},
	doi={10.1109/CEC.2008.4631255}}

@article{salimans2017es,
	title={Evolution Strategies as a Scalable Alternative to Reinforcement Learning}, 
	author={Tim Salimans and Jonathan Ho and Xi Chen and Szymon Sidor and Ilya Sutskever},
	year={2017},
	journal={arXiv:1703.03864},
	url={https://arxiv.org/abs/1703.03864}, 
}

@book{sutton2018reinforcement,
  title     = {Reinforcement Learning: An Introduction},
  author    = {Sutton, Richard S. and Barto, Andrew G.},
  year      = {2018},
  edition   = {2nd},
  publisher = {MIT Press},
  address   = {Cambridge, MA}
}

@misc{liu2024oat,
  title={OAT: A research-friendly framework for LLM online alignment},
  author={Liu, Zichen and Chen, Changyu and Wan, Xinyi and Du, Chao and Lee, Wee Sun and Lin, Min},
  year={2024}
}

@article{zeng2025simplerl,
  title={Simplerl-zoo: Investigating and taming zero reinforcement learning for open base models in the wild},
  author={Zeng, Weihao and Huang, Yuzhen and Liu, Qian and Liu, Wei and He, Keqing and Ma, Zejun and He, Junxian},
  journal={arXiv preprint arXiv:2503.18892},
  year={2025}
}

@article{hendrycks2021measuring,
  title={Measuring mathematical problem solving with the math dataset},
  author={Hendrycks, Dan and Burns, Collin and Kadavath, Saurav and Arora, Akul and Basart, Steven and Tang, Eric and Song, Dawn and Steinhardt, Jacob},
  journal={arXiv preprint arXiv:2103.03874},
  year={2021}
}

@article{yang2024qwen2,
  title={Qwen2. 5-math technical report: Toward mathematical expert model via self-improvement},
  author={Yang, An and Zhang, Beichen and Hui, Binyuan and Gao, Bofei and Yu, Bowen and Li, Chengpeng and Liu, Dayiheng and Tu, Jianhong and Zhou, Jingren and Lin, Junyang and others},
  journal={arXiv preprint arXiv:2409.12122},
  year={2024}
}

@article{hu2025open,
  title={Open-reasoner-zero: An open source approach to scaling up reinforcement learning on the base model},
  author={Hu, Jingcheng and Zhang, Yinmin and Han, Qi and Jiang, Daxin and Zhang, Xiangyu and Shum, Heung-Yeung},
  journal={arXiv preprint arXiv:2503.24290},
  year={2025}
}

@article{sheng2024hybridflow,
  title   = {HybridFlow: A Flexible and Efficient RLHF Framework},
  author  = {Guangming Sheng and Chi Zhang and Zilingfeng Ye and Xibin Wu and Wang Zhang and Ru Zhang and Yanghua Peng and Haibin Lin and Chuan Wu},
  year    = {2024},
  journal = {arXiv preprint arXiv: 2409.19256}
}

@article{shah2025comedy,
  title={A Comedy of Estimators: On KL Regularization in RL Training of LLMs},
  author={Shah, Vedant and Obando-Ceron, Johan and Jain, Vineet and Bartoldson, Brian and Kailkhura, Bhavya and Mittal, Sarthak and Berseth, Glen and Castro, Pablo Samuel and Bengio, Yoshua and Malkin, Nikolay and others},
  journal={arXiv preprint arXiv:2512.21852},
  year={2025}
}

@article{liu2025understanding,
  title={Understanding r1-zero-like training: A critical perspective},
  author={Liu, Zichen and Chen, Changyu and Li, Wenjun and Qi, Penghui and Pang, Tianyu and Du, Chao and Lee, Wee Sun and Lin, Min},
  journal={arXiv preprint arXiv:2503.20783},
  year={2025}
}

@article{yu2025dapo,
  title={Dapo: An open-source llm reinforcement learning system at scale},
  author={Yu, Qiying and Zhang, Zheng and Zhu, Ruofei and Yuan, Yufeng and Zuo, Xiaochen and Yue, Yu and Dai, Weinan and Fan, Tiantian and Liu, Gaohong and Liu, Lingjun and others},
  journal={arXiv preprint arXiv:2503.14476},
  year={2025}
}

@InProceedings{igel:cec03,
  author       = "Christian Igel",
  title        = "Neuroevolution for Reinforcement Learning Using
                  Evolution Strategies",
  booktitle    = "Proceedings of the 2003 Congress on Evolutionary
                  Computation",
  year         = 2003,
  pages        = "2588--2595",
}

@inproceedings{qiu:neurips24,
 author = {Qiu, Xin and Miikkulainen, Risto},
 booktitle = {Advances in Neural Information Processing Systems},
 doi = {10.52202/079017-4274},
 editor = {A. Globerson and L. Mackey and D. Belgrave and A. Fan and U. Paquet and J. Tomczak and C. Zhang},
 pages = {134507--134533},
 publisher = {Curran Associates, Inc.},
 title = {{Semantic Density}: Uncertainty Quantification for Large Language Models through Confidence Measurement in Semantic Space},
 volume = {37},
 year = {2024}
}

@Article{farquhar:nature24,
  author       = "Farquhar, Sebastian and Kossen, Jannik and Kuhn,
                  Lorenz and Gal, Yarin",
  journal      = "Nature",
  number       = 8017,
  pages        = "625--630",
  title        = "Detecting hallucinations in large language models
                  using semantic entropy",
  volume       = 630,
  year         = 2024
}

@article{such2017DeepNG,
	title={Deep Neuroevolution: Genetic Algorithms Are a Competitive Alternative for Training Deep Neural Networks for Reinforcement Learning},
	author={Felipe Petroski Such and Vashisht Madhavan and Edoardo Conti and Joel Lehman and Kenneth O. Stanley and Jeff Clune},
	journal={arXiv:1712.06567},
	year={2017},
	url={https://api.semanticscholar.org/CorpusID:5044808}
}

@inproceedings{chrabaszcz2018,
author = {Chrabaszcz, Patryk and Loshchilov, Ilya and Hutter, Frank},
title = {Back to basics: benchmarking canonical evolution strategies for playing Atari},
year = {2018},
isbn = {9780999241127},
publisher = {AAAI Press},
booktitle = {Proceedings of the 27th International Joint Conference on Artificial Intelligence},
pages = {1419–1426},
numpages = {8},
location = {Stockholm, Sweden},
series = {IJCAI'18}
}

@inproceedings{conti2018,
 author = {Conti, Edoardo and Madhavan, Vashisht and Petroski Such, Felipe and Lehman, Joel and Stanley, Kenneth and Clune, Jeff},
 booktitle = {Advances in Neural Information Processing Systems},
 editor = {S. Bengio and H. Wallach and H. Larochelle and K. Grauman and N. Cesa-Bianchi and R. Garnett},
 pages = {},
 publisher = {Curran Associates, Inc.},
 title = {Improving Exploration in Evolution Strategies for Deep Reinforcement Learning via a Population of Novelty-Seeking Agents},
 volume = {31},
 year = {2018}
}

@inproceedings{ouyang2022rlhf,
	author = {Ouyang, Long and Wu, Jeff and Jiang, Xu and Almeida, Diogo and Wainwright, Carroll L. and Mishkin, Pamela and Zhang, Chong and Agarwal, Sandhini and Slama, Katarina and Ray, Alex and Schulman, John and Hilton, Jacob and Kelton, Fraser and Miller, Luke and Simens, Maddie and Askell, Amanda and Welinder, Peter and Christiano, Paul and Leike, Jan and Lowe, Ryan},
	title = {Training language models to follow instructions with human feedback},
	year = {2022},
	isbn = {9781713871088},
	publisher = {Curran Associates Inc.},
	address = {Red Hook, NY, USA},
	booktitle = {Proceedings of the 36th International Conference on Neural Information Processing Systems},
	articleno = {2011},
	numpages = {15},
	location = {New Orleans, LA, USA},
	series = {NIPS '22}
}

@inproceedings{rafailov2023dpo,
	author = {Rafailov, Rafael and Sharma, Archit and Mitchell, Eric and Ermon, Stefano and Manning, Christopher D. and Finn, Chelsea},
	title = {Direct preference optimization: your language model is secretly a reward model},
	year = {2023},
	publisher = {Curran Associates Inc.},
	address = {Red Hook, NY, USA},
	booktitle = {Proceedings of the 37th International Conference on Neural Information Processing Systems},
	articleno = {2338},
	numpages = {14},
	location = {New Orleans, LA, USA},
	series = {NIPS '23}
}

@article{shao2024grpo,
	title={DeepSeekMath: Pushing the Limits of Mathematical Reasoning in Open Language Models}, 
	author={Zhihong Shao and Peiyi Wang and Qihao Zhu and Runxin Xu and Junxiao Song and Xiao Bi and Haowei Zhang and Mingchuan Zhang and Y. K. Li and Y. Wu and Daya Guo},
	year={2024},
	journal={arXiv:2402.03300},
	url={https://arxiv.org/abs/2402.03300}, 
}

@article{zhang2017essgd,
	title={On the Relationship Between the OpenAI Evolution Strategy and Stochastic Gradient Descent}, 
	author={Xingwen Zhang and Jeff Clune and Kenneth O. Stanley},
	year={2017},
	journal={arXiv:1712.06564},
	url={https://arxiv.org/abs/1712.06564}, 
}

@article{lorenc2025estransformer,
	title={Utilizing Evolution Strategies to Train Transformers in Reinforcement Learning}, 
	author={Matyáš Lorenc and Roman Neruda},
	year={2025},
	journal={arXiv:2501.13883},
	url={https://arxiv.org/abs/2501.13883}, 
}

@article{openai2024gpt4,
	title={{GPT}-4 Technical Report}, 
	author={Josh Achiam and others},
	year={2024},
	journal={arXiv:2303.08774},
}

@article{touvron2023llama2,
	title={Llama 2: Open Foundation and Fine-Tuned Chat Models}, 
	author={Hugo Touvron and others},
	year={2023},
	journal={arXiv:2307.09288},
}

@misc{llama3modelcard,
	title={Llama 3 Model Card},
	author={AI@Meta},
	year={2024},
	url = {https://github.com/meta-llama/llama3/blob/main/MODEL_CARD.md}
}

@article{jiang2024mixtral,
	title={Mixtral of Experts}, 
	author={Albert Q. Jiang and Alexandre Sablayrolles and Antoine Roux and Arthur Mensch and Blanche Savary and Chris Bamford and Devendra Singh Chaplot and Diego de las Casas and Emma Bou Hanna and Florian Bressand and Gianna Lengyel and Guillaume Bour and Guillaume Lample and Lélio Renard Lavaud and Lucile Saulnier and Marie-Anne Lachaux and Pierre Stock and Sandeep Subramanian and Sophia Yang and Szymon Antoniak and Teven Le Scao and Théophile Gervet and Thibaut Lavril and Thomas Wang and Timothée Lacroix and William El Sayed},
	year={2024},
	journal={arXiv:2401.04088},
}

@article{liu2024deepseek,
	title={Deepseek-v3 technical report},
	author={Liu, Aixin and Feng, Bei and Xue, Bing and Wang, Bingxuan and Wu, Bochao and Lu, Chengda and Zhao, Chenggang and Deng, Chengqi and Zhang, Chenyu and Ruan, Chong and others},
	journal={arXiv:2412.19437},
	year={2024}
}

@misc{claude4,
	title={Introducing {Claude} 4},
	author={Anthropic},
	year={2025},
	url = {https://www.anthropic.com/news/claude-4}
}

@misc{gemini25,
	title={Gemini 2.5: Pushing the Frontier with Advanced Reasoning, Multimodality, Long Context, and Next Generation Agentic Capabilities.},
	author={Google},
	year={2025},
	url = {https://storage.googleapis.com/deepmind-media/gemini/gemini_v2_5_report.pdf}
}

@article{singhal2023expertlevel,
	title={Towards Expert-Level Medical Question Answering with Large Language Models}, 
	author={Karan Singhal and Tao Tu and Juraj Gottweis and Rory Sayres and Ellery Wulczyn and Le Hou and Kevin Clark and Stephen Pfohl and Heather Cole-Lewis and Darlene Neal and Mike Schaekermann and Amy Wang and Mohamed Amin and Sami Lachgar and Philip Mansfield and Sushant Prakash and Bradley Green and Ewa Dominowska and Blaise Aguera y Arcas and Nenad Tomasev and Yun Liu and Renee Wong and Christopher Semturs and S. Sara Mahdavi and Joelle Barral and Dale Webster and Greg S. Corrado and Yossi Matias and Shekoofeh Azizi and Alan Karthikesalingam and Vivek Natarajan},
	year={2023},
	journal={arXiv:2305.09617},
}

@article{wu2023bloomberggpt,
	title={BloombergGPT: A Large Language Model for Finance}, 
	author={Shijie Wu and Ozan Irsoy and Steven Lu and Vadim Dabravolski and Mark Dredze and Sebastian Gehrmann and Prabhanjan Kambadur and David Rosenberg and Gideon Mann},
	year={2023},
	journal={arXiv:2303.17564},
}

@article{roziere2024code,
	title={Code Llama: Open Foundation Models for Code}, 
	author={Baptiste Rozière and Jonas Gehring and Fabian Gloeckle and Sten Sootla and Itai Gat and Xiaoqing Ellen Tan and Yossi Adi and Jingyu Liu and Romain Sauvestre and Tal Remez and Jérémy Rapin and Artyom Kozhevnikov and Ivan Evtimov and Joanna Bitton and Manish Bhatt and Cristian Canton Ferrer and Aaron Grattafiori and Wenhan Xiong and Alexandre Défossez and Jade Copet and Faisal Azhar and Hugo Touvron and Louis Martin and Nicolas Usunier and Thomas Scialom and Gabriel Synnaeve},
	year={2024},
	journal={arXiv:2308.12950},
}

@article{Romera-Paredes2024math,
	Author = {Romera-Paredes, Bernardino and Barekatain, Mohammadamin and Novikov, Alexander and Balog, Matej and Kumar, M. Pawan and Dupont, Emilien and Ruiz, Francisco J. R. and Ellenberg, Jordan S. and Wang, Pengming and Fawzi, Omar and Kohli, Pushmeet and Fawzi, Alhussein},
	Da = {2024/01/01},
	Doi = {10.1038/s41586-023-06924-6},
	Id = {Romera-Paredes2024},
	Isbn = {1476-4687},
	Journal = {Nature},
	Number = {7995},
	Pages = {468--475},
	Title = {Mathematical discoveries from program search with large language models},
	Ty = {JOUR},
	Url = {https://doi.org/10.1038/s41586-023-06924-6},
	Volume = {625},
	Year = {2024}}

@article{latif2024score,
	title = {Fine-tuning ChatGPT for automatic scoring},
	journal = {Computers and Education: Artificial Intelligence},
	volume = {6},
	pages = {100210},
	year = {2024},
	issn = {2666-920X},
	doi = {https://doi.org/10.1016/j.caeai.2024.100210},
	url = {https://www.sciencedirect.com/science/article/pii/S2666920X24000110},
	author = {Ehsan Latif and Xiaoming Zhai}
}

@article{bai2022assistant,
	title={Training a Helpful and Harmless Assistant with Reinforcement Learning from Human Feedback}, 
	author={Yuntao Bai and Andy Jones and Kamal Ndousse and Amanda Askell and Anna Chen and Nova DasSarma and Dawn Drain and Stanislav Fort and Deep Ganguli and Tom Henighan and Nicholas Joseph and Saurav Kadavath and Jackson Kernion and Tom Conerly and Sheer El-Showk and Nelson Elhage and Zac Hatfield-Dodds and Danny Hernandez and Tristan Hume and Scott Johnston and Shauna Kravec and Liane Lovitt and Neel Nanda and Catherine Olsson and Dario Amodei and Tom Brown and Jack Clark and Sam McCandlish and Chris Olah and Ben Mann and Jared Kaplan},
	year={2022},
	journal={arXiv:2204.05862},
	url={https://arxiv.org/abs/2204.05862}, 
}

@article{deepseekai2025deepseekr1,
	title={DeepSeek-R1: Incentivizing Reasoning Capability in LLMs via Reinforcement Learning}, 
	author={Daya Guo and others},
	year={2025},
	journal={arXiv:2501.12948},
	url={https://arxiv.org/abs/2501.12948}, 
}

@article{srivastava2025rlsurvey,
	title={A Technical Survey of Reinforcement Learning Techniques for Large Language Models}, 
	author={Saksham Sahai Srivastava and Vaneet Aggarwal},
	year={2025},
	journal={arXiv:2507.04136},
	url={https://arxiv.org/abs/2507.04136}, 
}

@inproceedings{choshen2020instablerl,
	title={On the Weaknesses of Reinforcement Learning for Neural Machine Translation},
	author={Leshem Choshen and Lior Fox and Zohar Aizenbud and Omri Abend},
	booktitle={International Conference on Learning Representations},
	year={2020},
	url={https://openreview.net/forum?id=H1eCw3EKvH}
}

@inproceedings{zhong2025dpo,
	title={{DPO} Meets {PPO}: Reinforced Token Optimization for {RLHF}},
	author={Han Zhong and Zikang Shan and Guhao Feng and Wei Xiong and Xinle Cheng and Li Zhao and Di He and Jiang Bian and Liwei Wang},
	booktitle={Forty-second International Conference on Machine Learning},
	year={2025},
	url={https://openreview.net/forum?id=IfWKVF6LfY}
}

@conference{vemula2019contrast,
	author = {Anirudh Vemula and Wen Sun and J. Andrew Bagnell},
	title = {Contrasting Exploration in Parameter and Action Space: A Zeroth Order Optimization Perspective},
	booktitle = {Proceedings of 22nd International Conference on Artificial Intelligence and Statistics (AISTATS '19)},
	year = {2019},
	month = {March},
}

@article{zhang2025lessons,
	title={The Lessons of Developing Process Reward Models in Mathematical Reasoning}, 
	author={Zhenru Zhang and Chujie Zheng and Yangzhen Wu and Beichen Zhang and Runji Lin and Bowen Yu and Dayiheng Liu and Jingren Zhou and Junyang Lin},
	year={2025},
	journal={arXiv:2501.07301},
	url={https://arxiv.org/abs/2501.07301}, 
}

@inproceedings{song2025prmbench,
	title = "{PRMB}ench: A Fine-grained and Challenging Benchmark for Process-Level Reward Models",
	author = "Song, Mingyang  and
	Su, Zhaochen  and
	Qu, Xiaoye  and
	Zhou, Jiawei  and
	Cheng, Yu",
	editor = "Che, Wanxiang  and
	Nabende, Joyce  and
	Shutova, Ekaterina  and
	Pilehvar, Mohammad Taher",
	booktitle = "Proceedings of the 63rd Annual Meeting of the Association for Computational Linguistics (Volume 1: Long Papers)",
	month = jul,
	year = "2025",
	address = "Vienna, Austria",
	publisher = "Association for Computational Linguistics",
	url = "https://aclanthology.org/2025.acl-long.1230/",
	doi = "10.18653/v1/2025.acl-long.1230",
	pages = "25299--25346",
	ISBN = "979-8-89176-251-0"
}

@inproceedings{gandhi2025cognitive,
	title={Cognitive Behaviors that Enable Self-Improving Reasoners, or, Four Habits of Highly Effective {ST}aRs},
	author={Kanishk Gandhi and Ayush K Chakravarthy and Anikait Singh and Nathan Lile and Noah Goodman},
	booktitle={Second Conference on Language Modeling},
	year={2025},
	url={https://openreview.net/forum?id=QGJ9ttXLTy}
}

@article{denison2024tampering,
	title={Sycophancy to Subterfuge: Investigating Reward-Tampering in Large Language Models}, 
	author={Carson Denison and Monte MacDiarmid and Fazl Barez and David Duvenaud and Shauna Kravec and Samuel Marks and Nicholas Schiefer and Ryan Soklaski and Alex Tamkin and Jared Kaplan and Buck Shlegeris and Samuel R. Bowman and Ethan Perez and Evan Hubinger},
	year={2024},
	journal={arXiv:2406.10162},
	url={https://arxiv.org/abs/2406.10162}, 
}

@article{fu2025shaping,
	title={Reward Shaping to Mitigate Reward Hacking in {RLHF}}, 
	author={Jiayi Fu and Xuandong Zhao and Chengyuan Yao and Heng Wang and Qi Han and Yanghua Xiao},
	year={2025},
	journal={arXiv:2502.18770},
	url={https://arxiv.org/abs/2502.18770}, 
}

@inproceedings{jia2025need,
	title={Do We Need to Verify Step by Step? Rethinking Process Supervision from a Theoretical Perspective},
	author={Zeyu Jia and Alexander Rakhlin and Tengyang Xie},
	booktitle={Forty-second International Conference on Machine Learning},
	year={2025},
	url={https://openreview.net/forum?id=4BfaPHfhJ0}
}

@inproceedings{lehman2018esmore,
	author = {Lehman, Joel and Chen, Jay and Clune, Jeff and Stanley, Kenneth O.},
	title = {ES is more than just a traditional finite-difference approximator},
	year = {2018},
	isbn = {9781450356183},
	publisher = {Association for Computing Machinery},
	address = {New York, NY, USA},
	url = {https://doi.org/10.1145/3205455.3205474},
	doi = {10.1145/3205455.3205474},
	booktitle = {Proceedings of the Genetic and Evolutionary Computation Conference},
	pages = {450–457},
	numpages = {8},
	location = {Kyoto, Japan},
	series = {GECCO '18}
}

@inproceedings{huang2025evolution,
	title = "When Evolution Strategy Meets Language Models Tuning",
	author = "Huang, Bo  and
	Jiang, Yuxin  and
	Chen, Mingyang  and
	Wang, Yi  and
	Chen, Hongyang  and
	Wang, Wei",
	editor = "Rambow, Owen  and
	Wanner, Leo  and
	Apidianaki, Marianna  and
	Al-Khalifa, Hend  and
	Eugenio, Barbara Di  and
	Schockaert, Steven",
	booktitle = "Proceedings of the 31st International Conference on Computational Linguistics",
	month = jan,
	year = "2025",
	address = "Abu Dhabi, UAE",
	publisher = "Association for Computational Linguistics",
	url = "https://aclanthology.org/2025.coling-main.357/",
	pages = "5333--5344"
}

@article{uesato2022math,
	title={Solving math word problems with process- and outcome-based feedback}, 
	author={Jonathan Uesato and Nate Kushman and Ramana Kumar and Francis Song and Noah Siegel and Lisa Wang and Antonia Creswell and Geoffrey Irving and Irina Higgins},
	year={2022},
	journal={arXiv:2211.14275},
	url={https://arxiv.org/abs/2211.14275}, 
}

@inproceedings{toledano2022mt5,
	title = {Fine-tuning mT5-based Transformer via CMA-ES for Sentiment Analysis},
	author = {Orlando Grabiel Toledano-López and Julio Madera and Hector González and Alfredo Simón-Cuevas and Thomas Demeester and Erik Mannens},
	year = {2022},
	url = {http://ceur-ws.org/Vol-3202/restmex-paper12.pdf},
	researchr = {https://researchr.org/publication/Toledano-LopezM22-0},
	cites = {0},
	citedby = {0},
	booktitle = {Proceedings of the Iberian Languages Evaluation Forum (IberLEF 2022) co-located with the Conference of the Spanish Society for Natural Language Processing (SEPLN 2022), A Coruña, Spain, September 20, 2022},
	editor = {Manuel Montes-y-Gómez and Julio Gonzalo and Francisco Rangel and Marco Casavantes and Miguel Ángel Álvare Carmona and Gemma Bel Enguix and Hugo Jair Escalante and Larissa A. de Freitas and Antonio Miranda-Escalada and Francisco J. Rodríguez-Sánchez and Aiala Rosá and Marco Antonio Sobrevilla Cabezudo and Mariona Taulé and Rafael Valencia-García},
	volume = {3202},
	series = {CEUR Workshop Proceedings},
	publisher = {CEUR-WS.org},
}

@article{jin2024lowrank,
	author = {Jin, Feihu and Liu, Yifan and Tan, Ying},
	title = {Derivative-Free Optimization for Low-Rank Adaptation in Large Language Models},
	year = {2024},
	issue_date = {2024},
	publisher = {IEEE Press},
	volume = {32},
	issn = {2329-9290},
	url = {https://doi.org/10.1109/TASLP.2024.3477330},
	doi = {10.1109/TASLP.2024.3477330},
	journal = {IEEE/ACM Trans. Audio, Speech and Lang. Proc.},
	month = oct,
	pages = {4607–4616},
	numpages = {10}
}

@article{chen2025loss,
	title={Understanding Pre-training and Fine-tuning from Loss Landscape Perspectives}, 
	author={Huanran Chen and Yinpeng Dong and Zeming Wei and Yao Huang and Yichi Zhang and Hang Su and Jun Zhu},
	year={2025},
	journal={arXiv:2505.17646},
	url={https://arxiv.org/abs/2505.17646}, 
}

@inproceedings{malladi2023mezo,
	author = {Malladi, Sadhika and Gao, Tianyu and Nichani, Eshaan and Damian, Alex and Lee, Jason D and Chen, Danqi and Arora, Sanjeev},
	booktitle = {Advances in Neural Information Processing Systems},
	editor = {A. Oh and T. Naumann and A. Globerson and K. Saenko and M. Hardt and S. Levine},
	pages = {53038--53075},
	publisher = {Curran Associates, Inc.},
	title = {Fine-Tuning Language Models with Just Forward Passes},
	url = {https://proceedings.neurips.cc/paper_files/paper/2023/file/a627810151be4d13f907ac898ff7e948-Paper-Conference.pdf},
	volume = {36},
	year = {2023}
}

@article{li2024numinamath,
  title={Numinamath: The largest public dataset in ai4maths with 860k pairs of competition math problems and solutions},
  author={Li, Jia and Beeching, Edward and Tunstall, Lewis and Lipkin, Ben and Soletskyi, Roman and Huang, Shengyi and Rasul, Kashif and Yu, Longhui and Jiang, Albert Q and Shen, Ziju and others},
  journal={Hugging Face repository},
  volume={13},
  number={9},
  pages={9},
  year={2024}
}

@article{lewkowycz2022solving,
  title={Solving quantitative reasoning problems with language models},
  author={Lewkowycz, Aitor and Andreassen, Anders and Dohan, David and Dyer, Ethan and Michalewski, Henryk and Ramasesh, Vinay and Slone, Ambrose and Anil, Cem and Schlag, Imanol and Gutman-Solo, Theo and others},
  journal={Advances in neural information processing systems},
  volume={35},
  pages={3843--3857},
  year={2022}
}

@inproceedings{he2024olympiadbench,
  title={Olympiadbench: A challenging benchmark for promoting agi with olympiad-level bilingual multimodal scientific problems},
  author={He, Chaoqun and Luo, Renjie and Bai, Yuzhuo and Hu, Shengding and Thai, Zhen and Shen, Junhao and Hu, Jinyi and Han, Xu and Huang, Yujie and Zhang, Yuxiang and others},
  booktitle={Proceedings of the 62nd Annual Meeting of the Association for Computational Linguistics (Volume 1: Long Papers)},
  pages={3828--3850},
  year={2024}
}

@book{schwefel1977es,
	author = {Schwefel, Hans-Paul},
	year = {1977},
	month = {01},
	pages = {},
	title = {Numerische Optimierung von Computermodellen mittels der Evo-lutionsstrategie},
	volume = {26},
	isbn = {9783764308766},
	journal = {Interdisciplinary Systems Research},
	doi = {10.1007/978-3-0348-5927-1}
}

@book{rechenberg1973es,
	title={Evolutionsstrategie: Optimierung technischer Systeme nach Prinzipien der biologischen Evolution},
	author={Rechenberg, I.},
	isbn={9783772803741},
	lccn={74320689},
	series={Problemata (Stuttgart)},
	url={https://books.google.com/books?id=-WAQAQAAMAAJ},
	year={1973},
	publisher={Frommann-Holzboog}
}

@ARTICLE{hansen2001cmaes,
	author={Hansen, Nikolaus and Ostermeier, Andreas},
	journal={Evolutionary Computation}, 
	title={Completely Derandomized Self-Adaptation in Evolution Strategies}, 
	year={2001},
	volume={9},
	number={2},
	pages={159-195},
	doi={10.1162/106365601750190398}}

@inproceedings{risi2019gaworld,
	author = {Risi, Sebastian and Stanley, Kenneth O.},
	title = {Deep neuroevolution of recurrent and discrete world models},
	year = {2019},
	isbn = {9781450361118},
	publisher = {Association for Computing Machinery},
	address = {New York, NY, USA},
	url = {https://doi.org/10.1145/3321707.3321817},
	doi = {10.1145/3321707.3321817},
	booktitle = {Proceedings of the Genetic and Evolutionary Computation Conference},
	pages = {456–462},
	numpages = {7},
	location = {Prague, Czech Republic},
	series = {GECCO '19}
}

@article{wang2025eallm,
	author = {Chao Wang  and Jiaxuan Zhao  and Licheng Jiao  and Lingling Li  and Fang Liu  and Shuyuan Yang },
	title = {When Large Language Models Meet Evolutionary Algorithms: Potential Enhancements and Challenges},
	journal = {Research},
	volume = {8},
	number = {},
	pages = {0646},
	year = {2025},
	doi = {10.34133/research.0646},
	URL = {https://spj.science.org/doi/abs/10.34133/research.0646},
	eprint = {https://spj.science.org/doi/pdf/10.34133/research.0646}}

@ARTICLE{xingyu2025eallm,
	author={Wu, Xingyu and Wu, Sheng-Hao and Wu, Jibin and Feng, Liang and Tan, Kay Chen},
	journal={IEEE Transactions on Evolutionary Computation}, 
	title={Evolutionary Computation in the Era of Large Language Model: Survey and Roadmap}, 
	year={2025},
	volume={29},
	number={2},
	pages={534-554},
	doi={10.1109/TEVC.2024.3506731}}

@inproceedings{guo2024connecting,
	title={Connecting Large Language Models with Evolutionary Algorithms Yields Powerful Prompt Optimizers},
	author={Qingyan Guo and Rui Wang and Junliang Guo and Bei Li and Kaitao Song and Xu Tan and Guoqing Liu and Jiang Bian and Yujiu Yang},
	booktitle={The Twelfth International Conference on Learning Representations},
	year={2024},
	url={https://openreview.net/forum?id=ZG3RaNIsO8}
}

@InProceedings{sun2022bbt,
	title = 	 {Black-Box Tuning for Language-Model-as-a-Service},
	author =       {Sun, Tianxiang and Shao, Yunfan and Qian, Hong and Huang, Xuanjing and Qiu, Xipeng},
	booktitle = 	 {Proceedings of the 39th International Conference on Machine Learning},
	pages = 	 {20841--20855},
	year = 	 {2022},
	editor = 	 {Chaudhuri, Kamalika and Jegelka, Stefanie and Song, Le and Szepesvari, Csaba and Niu, Gang and Sabato, Sivan},
	volume = 	 {162},
	series = 	 {Proceedings of Machine Learning Research},
	month = 	 {17--23 Jul},
	publisher =    {PMLR},
	url = 	 {https://proceedings.mlr.press/v162/sun22e.html}
}

@inproceedings{sun2022bbtv2,
	title = "{BBT}v2: Towards a Gradient-Free Future with Large Language Models",
	author = "Sun, Tianxiang  and
	He, Zhengfu  and
	Qian, Hong  and
	Zhou, Yunhua  and
	Huang, Xuanjing  and
	Qiu, Xipeng",
	editor = "Goldberg, Yoav  and
	Kozareva, Zornitsa  and
	Zhang, Yue",
	booktitle = "Proceedings of the 2022 Conference on Empirical Methods in Natural Language Processing",
	month = dec,
	year = "2022",
	address = "Abu Dhabi, United Arab Emirates",
	publisher = "Association for Computational Linguistics",
	url = "https://aclanthology.org/2022.emnlp-main.259/",
	doi = "10.18653/v1/2022.emnlp-main.259",
	pages = "3916--3930"
}

@inproceedings{zhao2023gaprompt,
	title     = {Genetic Prompt Search via Exploiting Language Model Probabilities},
	author    = {Zhao, Jiangjiang and Wang, Zhuoran and Yang, Fangchun},
	booktitle = {Proceedings of the Thirty-Second International Joint Conference on
	Artificial Intelligence, {IJCAI-23}},
	publisher = {International Joint Conferences on Artificial Intelligence Organization},
	editor    = {Edith Elkind},
	pages     = {5296--5305},
	year      = {2023},
	month     = {8},
	note      = {Main Track},
	doi       = {10.24963/ijcai.2023/588},
	url       = {https://doi.org/10.24963/ijcai.2023/588},
}

@article{meyerson2024lmx,
	author = {Meyerson, Elliot and Nelson, Mark J. and Bradley, Herbie and Gaier, Adam and Moradi, Arash and Hoover, Amy K. and Lehman, Joel},
	title = {Language Model Crossover: Variation through Few-Shot Prompting},
	year = {2024},
	issue_date = {December 2024},
	publisher = {Association for Computing Machinery},
	address = {New York, NY, USA},
	volume = {4},
	number = {4},
	url = {https://doi.org/10.1145/3694791},
	doi = {10.1145/3694791},
	journal = {ACM Trans. Evol. Learn. Optim.},
	month = nov,
	articleno = {27},
	numpages = {40}
}

@Inbook{lehman2024evolm,
	author="Lehman, Joel
	and Gordon, Jonathan
	and Jain, Shawn
	and Ndousse, Kamal
	and Yeh, Cathy
	and Stanley, Kenneth O.",
	title="Evolution Through Large Models",
	bookTitle="Handbook of Evolutionary Machine Learning",
	year="2024",
	publisher="Springer Nature Singapore",
	address="Singapore",
	pages="331--366",
	isbn="978-981-99-3814-8",
	doi="10.1007/978-981-99-3814-8_11",
	url="https://doi.org/10.1007/978-981-99-3814-8_11"
}

@article{novikov2025alphaevolve,
	title={{AlphaEvolve}: A coding agent for scientific and algorithmic discovery}, 
	author={Alexander Novikov and Ngân Vũ and Marvin Eisenberger and Emilien Dupont and Po-Sen Huang and Adam Zsolt Wagner and Sergey Shirobokov and Borislav Kozlovskii and Francisco J. R. Ruiz and Abbas Mehrabian and M. Pawan Kumar and Abigail See and Swarat Chaudhuri and George Holland and Alex Davies and Sebastian Nowozin and Pushmeet Kohli and Matej Balog},
	year={2025},
	journal={arXiv:2506.13131},
	url={https://arxiv.org/abs/2506.13131}, 
}

@inproceedings{du2024fusion,
	title = "Knowledge Fusion By Evolving Weights of Language Models",
	author = "Du, Guodong  and
	Li, Jing  and
	Liu, Hanting  and
	Jiang, Runhua  and
	Yu, Shuyang  and
	Guo, Yifei  and
	Goh, Sim Kuan  and
	Tang, Ho-Kin",
	editor = "Ku, Lun-Wei  and
	Martins, Andre  and
	Srikumar, Vivek",
	booktitle = "Findings of the Association for Computational Linguistics: ACL 2024",
	month = aug,
	year = "2024",
	address = "Bangkok, Thailand",
	publisher = "Association for Computational Linguistics",
	url = "https://aclanthology.org/2024.findings-acl.698/",
	doi = "10.18653/v1/2024.findings-acl.698",
	pages = "11727--11742"
}

@article{akiba2025merge,
	Author = {Akiba, Takuya and Shing, Makoto and Tang, Yujin and Sun, Qi and Ha, David},
	Da = {2025/02/01},
	Doi = {10.1038/s42256-024-00975-8},
	Id = {Akiba2025},
	Isbn = {2522-5839},
	Journal = {Nature Machine Intelligence},
	Number = {2},
	Pages = {195--204},
	Title = {Evolutionary optimization of model merging recipes},
	Ty = {JOUR},
	Url = {https://doi.org/10.1038/s42256-024-00975-8},
	Volume = {7},
	Year = {2025}}

@inproceedings{sanchez2024gatransformer,
	title = "How Well Can a Genetic Algorithm Fine-tune Transformer Encoders? A First Approach",
	author = "Sanchez Carmona, Vicente Ivan  and
	Jiang, Shanshan  and
	Dong, Bin",
	editor = "Tafreshi, Shabnam  and
	Akula, Arjun  and
	Sedoc, Jo{\~a}o  and
	Drozd, Aleksandr  and
	Rogers, Anna  and
	Rumshisky, Anna",
	booktitle = "Proceedings of the Fifth Workshop on Insights from Negative Results in NLP",
	month = jun,
	year = "2024",
	address = "Mexico City, Mexico",
	publisher = "Association for Computational Linguistics",
	url = "https://aclanthology.org/2024.insights-1.4/",
	doi = "10.18653/v1/2024.insights-1.4",
	pages = "25--33"
}

@article{liang2025deepseeknature,
	Author = {Guo, Daya and others},
	Da = {2025/09/01},
	Doi = {10.1038/s41586-025-09422-z},
	Id = {Guo2025},
	Isbn = {1476-4687},
	Journal = {Nature},
	Number = {8081},
	Pages = {633--638},
	Title = {DeepSeek-R1 incentivizes reasoning in LLMs through reinforcement learning},
	Ty = {JOUR},
	Url = {https://doi.org/10.1038/s41586-025-09422-z},
	Volume = {645},
	Year = {2025}}

@article{schulman2017ppo,
	title={Proximal Policy Optimization Algorithms}, 
	author={John Schulman and Filip Wolski and Prafulla Dhariwal and Alec Radford and Oleg Klimov},
	year={2017},
	journal={arXiv:1707.06347},
	url={https://arxiv.org/abs/1707.06347}, 
}

@InProceedings{gao2023scaling,
	title = 	 {Scaling Laws for Reward Model Overoptimization},
	author =       {Gao, Leo and Schulman, John and Hilton, Jacob},
	booktitle = 	 {Proceedings of the 40th International Conference on Machine Learning},
	pages = 	 {10835--10866},
	year = 	 {2023},
	editor = 	 {Krause, Andreas and Brunskill, Emma and Cho, Kyunghyun and Engelhardt, Barbara and Sabato, Sivan and Scarlett, Jonathan},
	volume = 	 {202},
	series = 	 {Proceedings of Machine Learning Research},
	month = 	 {23--29 Jul},
	publisher =    {PMLR},
	url = 	 {https://proceedings.mlr.press/v202/gao23h.html}
}

@article{sehnke2010parameter,
	title = {Parameter-exploring policy gradients},
	journal = {Neural Networks},
	volume = {23},
	number = {4},
	pages = {551-559},
	year = {2010},
	note = {The 18th International Conference on Artificial Neural Networks, ICANN 2008},
	issn = {0893-6080},
	doi = {https://doi.org/10.1016/j.neunet.2009.12.004},
	url = {https://www.sciencedirect.com/science/article/pii/S0893608009003220},
	author = {Frank Sehnke and Christian Osendorfer and Thomas Rückstieß and Alex Graves and Jan Peters and Jürgen Schmidhuber}
}

@article{thomas2010parameter,
	Author = {R{\"u}ckstie{\ss}, Thomas and Sehnke, Frank and Schaul, Tom and Wierstra, Daan and Sun, Yi and Schmidhuber, J{\"u}rgen},
	Da = {2010/03/01},
	Doi = {10.2478/s13230-010-0002-4},
	Id = {R{\"u}ckstie{\ss}2010},
	Isbn = {2080-9778},
	Journal = {Paladyn},
	Number = {1},
	Pages = {14--24},
	Title = {Exploring parameter space in reinforcement learning},
	Ty = {JOUR},
	Url = {https://doi.org/10.2478/s13230-010-0002-4},
	Volume = {1},
	Year = {2010}}

@inproceedings{plappert2018parameter,
	title={Parameter Space Noise for Exploration},
	author={Matthias Plappert and Rein Houthooft and Prafulla Dhariwal and Szymon Sidor and Richard Y. Chen and Xi Chen and Tamim Asfour and Pieter Abbeel and Marcin Andrychowicz},
	booktitle={International Conference on Learning Representations},
	year={2018},
	url={https://openreview.net/forum?id=ByBAl2eAZ},
}

@InProceedings{thomas2008state,
	author="R{\"u}ckstie{\ss}, Thomas
	and Felder, Martin
	and Schmidhuber, J{\"u}rgen",
	editor="Daelemans, Walter
	and Goethals, Bart
	and Morik, Katharina",
	title="State-Dependent Exploration for Policy Gradient Methods",
	booktitle="Machine Learning and Knowledge Discovery in Databases",
	year="2008",
	publisher="Springer Berlin Heidelberg",
	address="Berlin, Heidelberg",
	pages="234--249",
	isbn="978-3-540-87481-2"
}

@ARTICLE{spall1992spsa,
	author={Spall, J.C.},
	journal={IEEE Transactions on Automatic Control}, 
	title={Multivariate stochastic approximation using a simultaneous perturbation gradient approximation}, 
	year={1992},
	volume={37},
	number={3},
	pages={332-341},
	doi={10.1109/9.119632}}

@inproceedings{salimans2016gan,
	author = {Salimans, Tim and Goodfellow, Ian and Zaremba, Wojciech and Cheung, Vicki and Radford, Alec and Chen, Xi},
	title = {Improved techniques for training GANs},
	year = {2016},
	isbn = {9781510838819},
	publisher = {Curran Associates Inc.},
	address = {Red Hook, NY, USA},
	booktitle = {Proceedings of the 30th International Conference on Neural Information Processing Systems},
	pages = {2234–2242},
	numpages = {9},
	location = {Barcelona, Spain},
	series = {NIPS'16}
}

@inproceedings{kingma2015adam,
	author       = {Diederik P. Kingma and
	Jimmy Ba},
	editor       = {Yoshua Bengio and
	Yann LeCun},
	title        = {Adam: {A} Method for Stochastic Optimization},
	booktitle    = {3rd International Conference on Learning Representations, {ICLR} 2015,
	San Diego, CA, USA, May 7-9, 2015, Conference Track Proceedings},
	year         = {2015},
	url          = {http://arxiv.org/abs/1412.6980},
	bibsource    = {dblp computer science bibliography, https://dblp.org}
}

@inproceedings{aghajanyan2021intrinsic,
	title = "Intrinsic Dimensionality Explains the Effectiveness of Language Model Fine-Tuning",
	author = "Aghajanyan, Armen  and
	Gupta, Sonal  and
	Zettlemoyer, Luke",
	editor = "Zong, Chengqing  and
	Xia, Fei  and
	Li, Wenjie  and
	Navigli, Roberto",
	booktitle = "Proceedings of the 59th Annual Meeting of the Association for Computational Linguistics and the 11th International Joint Conference on Natural Language Processing (Volume 1: Long Papers)",
	month = aug,
	year = "2021",
	address = "Online",
	publisher = "Association for Computational Linguistics",
	url = "https://aclanthology.org/2021.acl-long.568/",
	doi = "10.18653/v1/2021.acl-long.568",
	pages = "7319--7328"
}

@misc{mucci2023super,
	title={What is artificial superintelligence?},
	author={Tim Mucci and Cole Stryker},
	year={2023},
	url={https://www.ibm.com/think/topics/artificial-superintelligence},
}

@misc{
liu2025sparse,
title={Sparse Me{ZO}: Less Parameters for Better Performance in Zeroth-Order {LLM} Fine-Tuning},
author={Yong Liu and Zirui Zhu and Chaoyu Gong and Minhao Cheng and Cho-Jui Hsieh and Yang You},
year={2025},
url={https://openreview.net/forum?id=4Kw4KAoVnx}
}

@misc{gandhi2024stream,
      title={Stream of Search (SoS): Learning to Search in Language}, 
      author={Kanishk Gandhi and Denise Lee and Gabriel Grand and Muxin Liu and Winson Cheng and Archit Sharma and Noah D. Goodman},
      year={2024},
      eprint={2404.03683},
      archivePrefix={arXiv},
      primaryClass={cs.LG},
      url={https://arxiv.org/abs/2404.03683}, 
}

@article{chollet2024arc,
  title={Arc prize 2024: Technical report},
  author={Chollet, Francois and Knoop, Mike and Kamradt, Gregory and Landers, Bryan},
  journal={arXiv preprint arXiv:2412.04604},
  year={2024}
}

@misc{reasoninggym,
      title={REASONING GYM: Reasoning Environments for Reinforcement Learning with Verifiable Rewards}, 
      author={Zafir Stojanovski and Oliver Stanley and Joe Sharratt and Richard Jones and Abdulhakeem Adefioye and Jean Kaddour and Andreas Köpf},
      year={2025},
      eprint={2505.24760},
      archivePrefix={arXiv},
      primaryClass={cs.LG},
      url={https://arxiv.org/abs/2505.24760}, 
}

@misc{arc_rl_report2025,
  title        = {ARC-RL: Reinforcement Learning for ARC-AGI},
  author       = {Priyanka Ranke},
  year         = {2025},
  howpublished = {\url{https://github.com/priyankaranke/arc_rl/blob/main/report.pdf}},
  note         = {Accessed: 2026-01-27}
}

@misc{chollet2019measureintelligence,
      title={On the Measure of Intelligence}, 
      author={François Chollet},
      year={2019},
      eprint={1911.01547},
      archivePrefix={arXiv},
      primaryClass={cs.AI},
      url={https://arxiv.org/abs/1911.01547}, 
}

@misc{gan2026neuralthicketsdiversetask,
      title={Neural Thickets: Diverse Task Experts Are Dense Around Pretrained Weights}, 
      author={Yulu Gan and Phillip Isola},
      year={2026},
      eprint={2603.12228},
      archivePrefix={arXiv},
      primaryClass={cs.LG},
      url={https://arxiv.org/abs/2603.12228}, 
}

@inproceedings{Narayanan2021megatron,
author = {Narayanan, Deepak and Shoeybi, Mohammad and Casper, Jared and LeGresley, Patrick and Patwary, Mostofa and Korthikanti, Vijay and Vainbrand, Dmitri and Kashinkunti, Prethvi and Bernauer, Julie and Catanzaro, Bryan and Phanishayee, Amar and Zaharia, Matei},
title = {Efficient large-scale language model training on GPU clusters using megatron-LM},
year = {2021},
isbn = {9781450384421},
publisher = {Association for Computing Machinery},
address = {New York, NY, USA},
url = {https://doi.org/10.1145/3458817.3476209},
doi = {10.1145/3458817.3476209},
booktitle = {Proceedings of the International Conference for High Performance Computing, Networking, Storage and Analysis},
articleno = {58},
numpages = {15},
location = {St. Louis, Missouri},
series = {SC '21}
}
